\documentclass[reqno]{amsart}[standalone]
\usepackage[T1]{fontenc}
\usepackage[ruled,linesnumbered]{algorithm2e}
\usepackage{amsfonts}
\usepackage{amsmath}
\usepackage{amssymb}
\usepackage{amsthm}
\usepackage[toc,page]{appendix}
\usepackage{bm}
\usepackage{booktabs}
\usepackage{pifont}
\usepackage{url}
\usepackage{comment}
\usepackage{multirow}
\usepackage{glossaries}
\usepackage{graphicx}
\usepackage{mathtools}
\usepackage[inline]{enumitem}
\usepackage{cases}
\usepackage{hyperref}
\usepackage{rotating}
\usepackage{subcaption}
\usepackage[table]{xcolor}
\usepackage{float}
\usepackage{siunitx}
\usepackage[top=1in, bottom=1.25in, left=1.25in, right=1.25in]{geometry}

\usepackage{todonotes} 
\usepackage{mathrsfs}
\usepackage[foot]{amsaddr}
\usepackage{caption}
\usepackage{mathbbol}
\usepackage[giveninits=true]{biblatex}
\AtEveryBibitem{
    \clearfield{urlyear}
    \clearfield{urlmonth}
    \clearfield{month}
    \clearfield{issn}
    \clearfield{isbn}
     \iffieldundef{doi}
    {}
    {\clearfield{url}}%
}

\graphicspath{{images/}}

\theoremstyle{definition}
\newtheorem{definition}{Definition}[section]

\theoremstyle{remark}
\newtheorem{remark}[definition]{Remark}

\theoremstyle{remark}
\newtheorem{example}[definition]{Example}

\theoremstyle{plain}

\newtheorem{corollary}[definition]{Corollary}
\newtheorem{proposition}[definition]{Proposition}

\newcommand{\R}{\mathbb{R}}
\newcommand{\mP}{\mathbb{P}}
\newcommand{\mE}{\mathcal{E}}
\newcommand{\mV}{\mathcal{V}}
\newcommand{\mL}{\mathcal{L}}
\newcommand{\bs}{\boldsymbol}

\newcommand{\NN}[1]{\mathcal{NN}_\mathrm{#1}}

\newacronym{ae}{AE}{Autoencoder}
\newacronym{cfd}{CFD}{Computational Fluid Dynamics}
\newacronym{cnn}{CNN}{Convolutional Neural Network}
\newacronym{dl}{DL}{Deep Learning}
\newacronym{dof}{DoF}{Degree of Freedom}
\newacronym{elu}{ELU}{Exponential Linear Unit}
\newacronym{fd}{FD}{Finite Difference}
\newacronym{fe}{FE}{Finite Element}
\newacronym{fom}{FOM}{Full-Order Method}
\newacronym{fno}{FNO}{Fourier Neural Operator}
\newacronym{fv}{FV}{Finite Volume}
\newacronym{gca}{GCA-ROM}{Graph Convolutional Autoencoder}
\newacronym{gcn}{GCN}{Graph Convolutional Netork}
\newacronym{gdl}{GDL}{Geometric Deep Learning}
\newacronym{gfn}{GFN}{Graph Feedforward Network}
\newacronym{gnn}{GNN}{Graph Neural Network}
\newacronym{gno}{GNO}{Graph Neural Operator}
\newacronym{gpr}{GPR}{Gaussian Process Regression}
\newacronym{lbfgs}{L-BFGS}{Limited-memory BFGS}
\newacronym{ld}{LD}{Lid-driven Cavity}
\newacronym{ldgcn}{LD-GCN}{Latent Dynamics Graph Convolutional Network}
\newacronym{ldnet}{LDNet}{Latent Dynamics Network}
\newacronym{mse}{MSE}{Mean Squared Error}
\newacronym{mh}{MH}{Moving Hole Advection}
\newacronym{mor}{MOR}{Model Order Reduction}
\newacronym{nn}{NN}{Neural Network}
\newacronym{no}{NO}{Neural Operator}
\newacronym{node}{NODE}{Neural Ordinary Differential Equation}
\newacronym{nrmse}{NRMSE}{Normalized Root Mean Square Error}
\newacronym{ode}{ODE}{Ordinary Differential Equation}
\newacronym{pde}{PDE}{Partial Differential Equation}
\newacronym{pod}{POD}{Proper Orthogonal Decomposition}
\newacronym{rb}{RB}{Reduced Basis}
\newacronym{rbf}{RBF}{Radial Basis Function}
\newacronym{rnn}{RNN}{Recurrent Neural Network}
\newacronym{rom}{ROM}{Reduced Order Model}
\newacronym{sa}{SA}{Square Advection}
\newacronym{sciml}{SciML}{Scientific Machine Learning}
\newacronym{shred}{SHRED}{Shallow Recurrent Decoder}
\newacronym{sindy}{SINDy}{Sparse Identification of Nonlinear Dynamics}
\newacronym{uat}{UAT}{Universal Approximation Theorem}

\usepackage{tikz,xcolor}
\definecolor{lime}{HTML}{A6CE39}
\DeclareRobustCommand{\orcidicon}{%
	\begin{tikzpicture}
	\draw[lime, fill=lime] (0,0) 
	circle [radius=0.16] 
	node[white] {{\fontfamily{qag}\selectfont \tiny ID}};
	\draw[white, fill=white] (-0.075,0.095) 
	circle [radius=0.007];
	\end{tikzpicture}
	\hspace{-2mm}
}
\foreach \x in {A, ..., Z}{%
	\expandafter\xdef\csname orcid\x\endcsname{\noexpand\href{https://orcid.org/\csname orcidauthor\x\endcsname}{\noexpand\orcidicon}}
}

\begin{document}
\title[LD-GCN for MOR of parameterized time-dependent PDEs]{Latent Dynamics Graph Convolutional Networks for model order reduction of parameterized time-dependent PDEs}
\author{Lorenzo Tomada$^{1}$ \orcidA, Federico Pichi$^{1}$ \orcidB, Gianluigi Rozza$^{1}$ \orcidC}
\address{$^1$ mathLab, Mathematics Area, SISSA, via Bonomea 265, I-34136 Trieste, Italy}
\email{\texttt{\{ltomada, fpichi, grozza\}@sissa.it}}
\begin{abstract} \glspl{gnn} are emerging as powerful tools for nonlinear \gls{mor} of time-dependent parameterized \glspl{pde}.
However, existing methodologies struggle to combine geometric inductive biases with interpretable latent behavior, overlooking dynamics-driven features or disregarding spatial information.

In this work, we address this gap by introducing \gls{ldgcn}, a purely data-driven, encoder-free architecture that learns a global, low-dimensional representation of dynamical systems conditioned on external inputs and parameters.
The temporal evolution is modeled in the latent space and advanced through time-stepping, allowing for time-extrapolation, and the trajectories are consistently decoded onto geometrically parameterized domains using a \gls{gnn}.
Our framework enhances interpretability by enabling the analysis of the reduced dynamics and supporting zero-shot prediction through latent interpolation.

The methodology is mathematically validated via a universal approximation theorem for encoder-free architectures, and numerically tested on complex computational mechanics problems involving physical and geometric parameters, including the detection of bifurcating phenomena for Navier--Stokes equations.


\begin{center}
    \textbf{Code availability:} \url{https://github.com/lorenzotomada/ld-gcn-rom}
\end{center}
\end{abstract}
\maketitle
\glsresetall
\tableofcontents
\section{Introduction}
The study of parameterized time-dependent \glspl{pde} is of utmost importance in numerous fields, with relevant applications in engineering \cite{zhao_lesnets_2025,li_geometry-informed_2023}, physics \cite{li_geometric_2025, pfaff_learning_2021}, and medicine \cite{fedele_comprehensive_2023, suk_mesh_2024}.
Exploring their behavior across multiple parameter configurations using high-fidelity numerical methods (e.g., \gls{fe} and \gls{fv}) is often computationally prohibitive, especially in the real-time and many-query context.
To address this challenge, \glspl{rom} have been developed to significantly reduce computational costs while maintaining accurate predictions, enabling efficient and reliable simulations \cite{benner_model_2017}.

A widely adopted \gls{mor} technique is the \gls{rb} method \cite{hesthaven_certified_2016, quarteroni_reduced_2016, rozza_real_2024}, which is based on a linear expansion ansatz and a two-stage ``offline-online'' strategy to obtain interpretable approximations and efficient evaluations.
However, traditional approaches to building reduced bases, such as \gls{pod} or Greedy algorithms, are limited by the linearity assumption and perform poorly in the presence of complex nonlinear behaviors, i.e., when the Kolmogorov $n$-width decays slowly \cite{OhlbergerReducedBasisMethods2016,hesthaven_certified_2016}.

Recent advances in \gls{dl} introduced new computational paradigms, offering promising alternatives to overcome these limitations \cite{brunton_data-driven_2019,quarteroni_combining_2025}.
In the context of data-driven modeling of parameterized dynamical systems, two research directions that have gained particular attention in recent years are \glspl{gnn} \cite{hamilton_graph_2020} (more broadly, \gls{gdl} \cite{bronstein_geometric_2021}) and latent dynamics models \cite{bonneville_extrapolating_2025}.

\glspl{gnn} have shown remarkable success across multiple domains \cite{pfaff_learning_2021, brandstetter_geometric_2022} due to their ability to exploit the relational structures of graph-based data and preserve symmetries \cite{xu_equivariant_2024,horie_graph_2024}.
In the context of differential problems, they provide a natural framework for handling unstructured meshes and complex geometries \cite{bankestad_flexible_2024,suk_mesh_2024}.
In contrast, latent dynamics models \cite{conti_reduced_2023,champion_data-driven_2019} assume that the evolution of the system can be represented using a reduced set of coordinates \cite{bonneville_comprehensive_2024}, which are advanced in time using models such as \glspl{node} \cite{chen_neural_2018}, \gls{sindy} \cite{brunton_discovering_2016}, and \glspl{rnn} \cite{faraji_shallow_2025}, and then efficiently decoded to reconstruct the full dynamics in a compact and interpretable way.

From a \gls{rom} perspective, the construction of such latent coordinates can be seen as a generalization of classical linear \gls{rom} techniques \cite{bao_regularized_2020}, where dimensionality reduction is traditionally performed via projection onto low-dimensional linear subspaces \cite{hesthaven_certified_2016}.
Recent advances in machine learning have enabled the extension of these ideas to nonlinear manifolds, allowing for more expressive and data-driven representations of complex dynamical systems \cite{champion_data-driven_2019}.

Within this framework, \glspl{ae} have emerged as a widely adopted tool for nonlinear dimensionality reduction \cite{brunton_data-driven_2019}, including graph-based variants that explicitly account for the underlying mesh or connectivity \cite{pichi_graph_2024, morrison_gfn_2024}.
More recently, increasing attention has been devoted to the development of encoder-free latent models \cite{regazzoni_learning_2024,faraji_shallow_2025,gao_sparse_2025}.
While these architectures lack a native mechanism to handle problems with varying initial conditions, they yield more interpretable latent trajectories and better capture the intrinsic dynamical features of the system.

To the best of our knowledge, no existing approach combines encoder-free graph-based models with a global latent representation that captures the underlying dynamics.
Indeed, similar methodologies rely on compression mechanisms to introduce a global latent state \cite{han_predicting_2022,ma_learning_2022}.
Although this state is not graph-structured, it is still derived from graph operations, such as pooling and concatenation of node features.
As a result, these approaches do not fully decouple the intrinsic dynamical evolution from geometric information.
Moreover, architectures such as \gls{gca} \cite{pichi_graph_2024} and \gls{gfn} \cite{morrison_gfn_2024}, which employ dimensionality reduction and feedforward/graph autoencoders, lack intrinsic causality and are thus unsuitable for dynamical systems.
Other \gls{gnn}-based methods for time-dependent \glspl{pde} typically evolve the full states via discrete time-stepping \cite{pfaff_learning_2021,wurth_physics-informed_2024} or learn \glspl{ode} directly on grid points \cite{liu_graph_2025}.

The main contribution of this work is the introduction of \gls{ldgcn}, a novel data-driven encoder-free architecture for reduced order modeling that integrates latent dynamics with graph-based decoding, enabling interpretable, efficient and accurate reconstruction of complex dynamical systems at reduced computational cost.
Building on the two-branch approach of \gls{ldnet} \cite{regazzoni_learning_2024}, our architecture comprises: (i) a \gls{node} that models and recurrently evolves the latent dynamics, and (ii) a graph convolutional decoder, inspired by \gls{gca}, that reconstructs full-coordinate solutions on varying unstructured meshes.
This design enables causal modeling within an encoder-free framework, ensuring flexibility, generalization across complex geometries, and non-intrusive applicability.
We further investigate interpolation in the latent space and establish a simple error bound under Lipschitz continuity assumptions for the decoder.
Importantly, the interpretability and regularity of latent trajectories naturally promote accurate interpolation.
Finally, we mathematically justify the proposed framework by proving a \gls{uat} for encoder-free architectures, and numerically validate \gls{ldgcn} on computational mechanics problems with physical and geometric parameters.
These include an advection-diffusion problem, the lid-driven cavity benchmark, and a bifurcating Navier--Stokes system, for which the method extracts meaningful behavior of the symmetry-breaking phenomenon \cite{quaini_symmetry_2016, pichi_artificial_2023}.
The results show that \gls{ldgcn} provides geometrically and temporally consistent results, outperforming \gls{gca} while using $50\%$ fewer trainable parameters, and achieving a performance comparable to \gls{ldnet}.

This work is organized as follows.
After a brief review of the literature on \glspl{gnn} for parameterized, time-dependent \glspl{pde}, Section \ref{sec:methodology} introduces the notation used throughout the article, provides an overview of the fundamental concepts of \glspl{gnn} and presents in detail our main contribution, i.e., the \gls{ldgcn} architecture.
Finally, Section \ref{sec:results} shows the performance of the methodology on various benchmarks, while Section \ref{sec:conclusions_and_perspectives} summarizes our contributions and discusses future perspectives and developments.

\subsection{Related works}\label{subsec:related}
In recent years, several paradigms have been proposed and further extended to design \gls{rom} strategies for parameterized time-dependent \glspl{pde}, including \glspl{gnn} and latent dynamics models.
A foundational contribution has been introduced in \cite{pfaff_learning_2021} exploiting a \glspl{gnn} in an autoregressive fashion: given the solution $\bs u_{t_k}^{\mathcal G}$ at time $t_k$ defined on a graph $\mathcal G$, the network learns the time derivative $\text{GNN}(\bs u_{t_k}^{\mathcal G}, \mathcal G)$, allowing the solution to be updated as $\bs u_{t_{k+1}}^{\mathcal G}=\bs u_{t_k}^{\mathcal G}+\text{GNN}(\bs u_{t_k}^{\mathcal G}, \mathcal G)$.
Several variants of this approach have been proposed to handle physics-informed losses and parameterized contexts \cite{wurth_physics-informed_2024,zhang_combining_2025,franco_deep_2023}.

In addition, alternative paradigms have emerged for modeling time evolution on graph structures, for instance, by generalizing classical solvers \cite{brandstetter_message_2022,bouziani_structure-preserving_2024}, possibly emphasizing the preservation of physical laws and symmetries \cite{horie_graph_2024}.
Researchers have also focused on the integration of differentiable solvers into the \gls{mor} pipeline for both steady and unsteady problems \cite{belbute-peres_combining_2020, yavich_differentiable_2025}, as well as designing physics-enhanced architectures that do not rely on one-time stepping \cite{gao_physics-informed_2022,lin_enabling_2025}, for example to perform data-driven system identification \cite{hernandez_thermodynamics-informed_2024}.
Significant progress has been made with the introduction of graph \glspl{node} \cite{poli_graph_2021}, which model the time derivative of the system in a continuous manner using a \gls{gnn} and have inspired various other architectures \cite{liu_graph_2025, iakovlev_learning_2021}.

Important contributions have also been made within the \glspl{no} framework~\cite{kovachki_neural_2023}, which generalizes the \gls{mor} paradigm by learning maps between function spaces, for example, approximating solutions to \glspl{pde} that depend on spatial fields. 
Models such as \glspl{gno} \cite{li_neural_2020} and their variants \cite{lin_enabling_2025} have achieved strong results on parameterized \glspl{pde}, but often treat time merely as a parameter rather than explicitly modeling causality, with some notable exceptions \cite{xu_equivariant_2024, wang_latent_2025}.

In parallel, extensive work on the identification of dynamical systems \cite{quarteroni_combining_2025} has led to the development of tools such as \glspl{node} \cite{chen_neural_2018} and \gls{sindy} \cite{brunton_discovering_2016}, sometimes incorporating \glspl{gnn} \cite{hernandez_thermodynamics-informed_2024, poli_graph_2021} to capture relational dependencies.
A flourishing branch of research focuses directly on modeling dynamics in latent spaces \cite{liu_hierarchical_2022,oommen_learning_2022,duan_non-intrusive_2024,farenga_latent_2025,conti_reduced_2023}, in which case dimensionality reduction is often achieved through \gls{pod} or \glspl{ae} \cite{wang_model_2018,farenga_latent_2025,conti_reduced_2023,tomada_sparse_2025, longhi_latent_2026, fries_lasdi_2022}, occasionally even providing suitable error bounds~\cite{farenga_latent_2025, brivio_error_2024}.
In contrast, encoder-free architectures such as \gls{ldnet} \cite{regazzoni_learning_2024} and \gls{shred}~\cite{faraji_shallow_2025,gao_physics-informed_2022,tomasetto_reduced_2025} have gained attention for their ability to learn latent coordinates that capture key features of the system while substantially reducing the number of trainable parameters.
In the case of \gls{ldnet}, this is achieved through a two-branch architecture that completely decouples the dynamical evolution in the latent space from the reconstruction of the output at the desired spatial coordinates.

In line with our aim, recent works have begun exploring the integration of \glspl{gnn} with the modeling of dynamics.
Existing methodologies typically evolve the dynamics on a (potentially smaller) graph~\cite{liu_graph_2025,barwey_multiscale_2023}, discover a latent state that is not geometry-independent \cite{ma_learning_2022,han_predicting_2022}, or employ (graph) \glspl{node} \cite{liu_segno_2024,hsieh_galds_2025}.
Further investigations have explored attention-based modeling of a latent state obtained via a graph \gls{ae} \cite{han_predicting_2022}, as well as stochastic frameworks \cite{bergna_uncertainty_2025}.

Additionally, existing architectures that define a global latent state and use it to reconstruct the full-order solution typically target generative modeling \cite{simonovsky_graphvae_2018, valencia_learning_2025} rather than individual trajectory prediction, or they often overlook dynamical information by treating time as a parameter.
This is the case of \gls{gca} \cite{pichi_graph_2024} and \gls{gfn} \cite{morrison_gfn_2024}, which employ a parameter-to-latent state map and may struggle when dealing with complex systems.
Further developments aimed at decoupling dynamical and parametric information have shown promising results for in-time extrapolation; however, approaches that rely on tensor-train decomposition are still inherently linear \cite{chen_time_2025}.

Finally, efficiency considerations have long motivated the use of interpolation techniques in \gls{mor} and machine learning to construct surrogate maps and augment data \cite{rasmussen_gaussian_2005,khamlich_optimal_2025}, particularly in the context of non-intrusive, regression-based strategies such as \gls{pod}-\gls{nn} \cite{hesthaven_non-intrusive_2018} or general interpolation tasks \cite{bui-thanh_proper_2003, demo_non-intrusive_2019}.
In these approaches, interpolation is performed directly in a reduced or latent space, enabling efficient predictions for unseen parameter values. 

Our work addresses the limitations mentioned above by introducing \glspl{ldgcn} within a non-intrusive, purely data-driven framework, combining latent dynamics with graph-based decoding for efficient \gls{mor}.
\section{Methodology}\label{sec:methodology}
The development of \glspl{rom} for parameterized and time-dependent \glspl{pde} requires approaches that can simultaneously capture the intrinsic dynamics of the system and remain flexible enough to deal with complex geometries and unstructured meshes.
Classical projection-based methods may struggle in this regard, especially in presence of strong nonlinear behaviors and parametric geometries, motivating the use of data-driven techniques that combine advances in latent dynamics modeling and \gls{gdl}.
Latent dynamics models aim to represent the temporal evolution of a system in terms of a compact set of dynamical variables, thus reducing the dimensionality while retaining the interpretability of the trajectories.
\glspl{gnn}, on the other hand, provide a natural framework for handling unstructured discretizations by exploiting the connection of the computational mesh and embedding geometric inductive biases.
Taking these two paradigms together, we aim at offering a powerful setting for non-intrusive, data-driven model reduction of \glspl{pde}, where the dynamics can be learned in a low-dimensional space and decoded back onto complex domains.

\subsection{Formal problem statement} \label{subsec:problem_statement}

Consider the following time-dependent differential problem:
        \begin{equation}\label{eq:abstract}
        \begin{cases}
              \partial_t\bm u(\bm x, t) = \mathcal{F}(\bm u(\bm x,t), t; \bm\mu(t)), &\forall(\bm{x},t)\in \Omega\times (0, T],\\
              \bm u(\bm x, 0)=\bm{u}_0(\bm x), &\forall\bm x\in\Omega,
        \end{cases}
        \end{equation}
with $\Omega\subset\R^d$ the space domain, $T\in\R^+$ the final time, $\mP \ni \bm\mu:I\to\mathcal A$ a signal\footnote{When $\bm\mu\in\mP$ is constant in time it can be seen as parameter; in such case, with slight abuse of notation, we identify functions in $\mP$ with constants in $\R^{d_{\bm\mu}}$.} active in the time interval $I=[0, T]$, $\mathcal A\subset\R^{d_{\bm\mu}}$ the set of admissible signal values, $\mathbb U \ni \bm u:\Omega\times I\to\R^{d_{\bm u}}$ the unknown solution to \eqref{eq:abstract}, being $\mathbb U$ some suitable function space, and $\mathcal F$ a suitable differential operator.
We also assume that, for each signal $\bm\mu\in\mP$, a single corresponding initial condition $\bm u_0$ is considered.
Our goal is to predict the field $\bm u$ given a signal $\bm\mu\in\mP$.

To ensure the well-posedness of \eqref{eq:abstract}, we assume that the signal-to-solution map $\mathcal S:\mP\to\mathbb U$, which maps $\bm\mu$ to the corresponding solution $u(\bs{x}, t; \bs{\mu})$, is well-defined, and that it is consistent with the arrow of time; that is, $\bm u(\bm x, t; \bs{\mu})$ depends only on $\bm\mu(s)|_{s\in[0, t]}$ and not on $\bm\mu(s)|_{s\in(t, T]}$.
Under such assumptions, we are interested in learning $\mathcal{S}$ in a non-intrusive way, i.e., given a finite set of observations.

\begin{remark}
It would have been possible to write Equation \eqref{eq:abstract} as an autonomous system simply by defining $\tilde{\bm\mu}(t)=[\bm\mu,\,t]^T\in\R^{d_{\bm\mu}+1}$ and expressing $\mathcal F$ as a function of $\tilde{\bm\mu}$.
However, for clarity purposes, we write the dependence on time explicitly.
\end{remark}

In particular, we rely on high-fidelity simulations based on the \gls{fe} method \cite{quarteroni_numerical_2017} to generate a dataset for the supervised learning procedure.
To this end, we define a discrete time grid $I_h=\{t_0,\dots,t_{N_t}\}$ and a graph $\mathcal M(\mathcal{V}, \mathcal{E})$, where $\mathcal{V}$ denotes the set of grid points and $\mathcal{E}$ encodes the mesh connectivity.
We assume that the mesh connectivity is fixed, and that the set of mesh nodes contains $N_h$ points for each simulation\footnote{The dataset is obtained by selecting the values of the high-fidelity solution at each mesh vertex, independently of the polynomial order of the Lagrange spaces used for the \gls{fe} method.}.
We also define $\mP_h = \{\bm{\mu}_i : I_h \to \mathcal A\}_{i=1}^{N_{\bm\mu}}$ as a finite set of realizations of the input signal $\bm\mu$.
For each $\bm\mu\in\mP$, the corresponding full-order solution at the grid points $\bm u_h(t;\bm\mu)\in{\R^{N_h\times {d_{\bm u}}}}$ is computed for $(t,\bm\mu)\in I_h\times\mP_h$.

Then, the dataset for the non-intrusive learning task simply consists of the input-output pairs $\mathcal{D}_\text{train}=\{(t,\bm\mu),\, \bm{u}_h(t;\bm\mu)\}_{(t,\bm\mu)\in\mathcal{T}_\text{train}}$, where the training input set is given by $\mathcal{T}_\text{train}=I_{h,\text{train}}\times\mP_{h,\text{train}}$.
Here, $I_{h,\text{train}} = \{t_0,\dots,t_{N_{t,\text{train}}}\} \subset I_h$ denotes a subset of the available time instances, while $\mP_{h,\text{train}} \subset \mP_h$ is a subset of the parameter realizations.
The cardinalities satisfy $N_{t,\text{train}} < N_t$ and $|\mP_{h,\text{train}}| = N_{\bm{\mu},\text{train}} < N_{\bm{\mu}}$, allowing us to assess the model’s ability to extrapolate in time and/or generalize to previously unseen parameter values.

\subsection{Brief introduction to graph neural networks}\label{subsec:GNN}
Given a mesh $\mathcal M(\mathcal V, \mathcal E)$, we can interpret it as a simple, connected, and undirected graph $\mathcal G (\mathcal V, \mathcal E)$ with nodes $\mV$ and edges $\mE$.
Each node has associated features $\bm u$ that correspond to the evaluation of a set of state variables at the
vertices of the mesh.
Global information can be encapsulated in the matrix $\bm U=[\bm{u}_1\,|\, \dots\,|\,\bm{u}_{N_h}]^T\in\R^{N_h\times d_{\bm u}}$, with $N_h$ nodes in the graph (with no particular order), each with $d_{\bm u}$ features (scalar or vector solution field).
Furthermore, we define the \emph{adjacency matrix} $\bm A\in\R^{N_h\times N_h}$ to keep track of the connections among nodes, with entries $a_{ij}=1$ if $(v_i, v_j)\in\mE$, and $a_{ij}=0$ otherwise.
A more convenient way to represent $\bm{A}$ is to consider the \emph{adjacency list} $\mE$, where each element $e=(i,j)$ represents the existence of an edge between $v_i$ and $v_j$.

To design a \gls{gnn}, we need to define a set of optimizable operations acting on (potentially all) attributes of the graph itself.
Although the study of \glspl{gnn} is one of the most active areas of deep learning research in recent years \cite{bronstein_geometric_2021}, most \gls{gnn} architectures are built on the concepts of \emph{neural message passing} \cite{hamilton_graph_2020} and \emph{convolutional layers}\footnote{Not all convolutional layers can be formulated as message passing, and vice versa.}.
We therefore proceed by introducing these two mechanisms.

Assume that we are given a set of hidden embeddings $\{\bm h^{(k)}_v\in\R^{d_{k}}\}$, where for each node $v\in\mathcal V$, the vector $\bm h^{(k)}_v$ represents its embedding at the $k$-layer, obtained via suitable differentiable computations.
The core idea underlying message passing is, for each $v\in\mV$, to update $\bm h^{(k)}_v$ by aggregating the information that is propagated from the set of local neighbors $N(v)$ of $v$.
The procedure consists in three distinct phases: (i) message computation, (ii) aggregation, and (iii) update.

More in detail, for each node $v\in N(u)$ we compute a message via the function $\mathfrak{m}^{(k)}$ to be sent to its neighbors, obtaining
\begin{equation*}
    \bm m^{(k)}_v=\mathfrak{m}^{(k)}\left(\bm h^{(k-1)}_v\right),\quad\forall v\in\mathcal V.
\end{equation*}
For a given node $u\in\mV$, the incoming messages from its neighbors $N(u)$ are aggregated using a function $\mathfrak{a}^{(k)}$, yielding
\begin{equation*}
    \bm m_{N(u)}^{(k)}= \mathfrak{a}^{(k)}\left(\{\bm m^{(k)}_v,\;\forall v\in N(u)\}\right),
\end{equation*}
and the hidden embeddings are finally updated using the function $\mathfrak{u}^{(k)}$:
\begin{equation*}
    \bm h_u^{(k)}=\mathfrak{u}^{(k)}\left(\bm m_{N(u)}^{(k)}\right).
\end{equation*}

\begin{example}
A simple message function $\mathfrak{m}^{(k)}$ consists of multiplying the hidden embedding by a weight matrix $\bm W^{(k)}\in\R^{d_{k}\times d_{k-1}}$, i.e.,
\begin{equation*}
      \bm m^{(k)}_v=\bm W^{(k)}\bm h^{(k-1)}_v,\quad\forall v\in\mathcal V.
\end{equation*}
A typical example of $\mathfrak a^{(k)}$ is given by the normalized sum of neighbors' embeddings (to avoid overweighting contributions from high-degree nodes), and is expressed as $\bm m_{N(u)}^{(k)}=\sum_{v\in N(u)}\frac{\bm m^{(k)}_v}{|N(u)|}$.
A common choice for the update function $\mathfrak{u}^{(k)}$ is given by some nonlinear activation function $\bm h_u^{(k)}=\sigma\left(m_{N(u)}^{(k)}\right)$, following standard practice in \glspl{nn}.

Finally, assuming that we also employ a bias term $\bm{b}\in\R^{d_k}$ to improve the performance, a basic example of \gls{gnn} with $K$ layers can be obtained as
\begin{equation}\label{eq:gcn}
    \bm h_u^{(k)}=\sigma\left(\frac{1}{|N(u)|}\sum_{v \in N(u)}\bm W^{(k)}\bm h_v^{(k-1)} +\bm b^{(k)}\right), \quad\forall u \in\mathcal{V},\;\forall k=1,\dots, K.
\end{equation}

\end{example}

Among \glspl{gnn}, an important class of architectures is that of \glspl{gcn}, which are designed to generalize the notion of convolution to unstructured domains.
Indeed, standard \glspl{cnn} can only operate on structured computational meshes, limiting their applicability to the context of regular grids.
Similarly to \glspl{cnn}, \glspl{gcn} employ trainable filters to produce aggregated information, but they also benefit from permutation invariance property.

It is possible to distinguish between \emph{spectral} and \emph{spatial} convolutions \cite{hamilton_graph_2020}.
Spectral approaches typically build a suitable basis by spectral analysis of the graph Laplacian matrix.
While these methods are widely used (also as a result of their solid mathematical foundations), they require an eigen-decomposition to define the filters, making them unsuitable to challenging scenarios.
In contrast, spatial convolutions mirror the \gls{cnn} approach and are typically defined by means of message passing.
In light of these considerations, spatial convolutions on graphs offer a natural choice for data-driven modeling of complex and large-scale systems, for which we aim at preserving in-time causality by coupling them with dynamics-oriented strategies.

A notable example of \gls{mor} strategy in this setting is \gls{gca}, which employs a graph convolutional \gls{ae} and a parameter-to-latent map.
In its offline phase, given a solution value $\bm u(\bm\mu)$, dimensionality reduction is performed by means of an encoder, obtaining $\bm z(\bm\mu)$.
The map is then tasked with modeling the dependence of $\bm z(\bm\mu)$ on $\bm\mu$, and the decoder learns the field reconstruction as a function of the latent state.
In its online phase, the query value is mapped to the latent space and then decoded.
\gls{gca} has demonstrated strong performance across a variety of reduced-order modeling tasks; however, its formulation does not enforce causal temporal evolution, which limits its applicability to systems exhibiting complex dynamics.

\subsection{Latent Dynamics Networks}
\glspl{ldnet} are two-branch meshless architectures specifically designed for time-dependent systems.
The core underlying idea is that, given a signal $\bm{\mu}\in\mP$, the reconstruction of the solution field $\bm{u}(\bm{x},t;\bm{\mu})$, with $(\bm{x},t)\in\Omega\times I$, is achieved by first evolving the system dynamics in a latent space to obtain a low-dimensional state $\bm{s}(t)$, and subsequently decoding $\bm{s}(t)$ back to the physical space.

More precisely, the first sub-network $\NN{dyn}: I \times \R^n \times \R^{d_{\bm{\mu}}} \to \R^n$ is a fully connected feedforward neural network that acts as a \gls{node} in the latent space $\R^n$.
Given a time $t\in I$, the latent state $\bm{s}(t)\in\R^n$, and the input signal $\bm{\mu}(t)\in\R^{d_{\bm\mu}}$, the network $\NN{dyn}$ predicts the time derivative $\frac{d}{dt}\bm{s}(t)\in\R^n$.
This defines the following system of \glspl{ode} governing the latent dynamics:
\begin{equation}\label{eq:latent_ODE_1}
    \frac{d}{dt}\bm{s}(t)=\NN{dyn}\bigl(t,\bm{s}(t),\bm{\mu}(t)\bigr),\quad t\in(0,T],
\end{equation}
which is then integrated in time.
The initial condition $\bm{s}(0)=\bm{0}$ is chosen without loss of generality as a consequence of the assumption made in Subsection \ref{subsec:problem_statement} that only one trajectory is considered for each instance of $\bm\mu\in\mP$ \cite{regazzoni_learning_2024}.

The second sub-network, $\NN{dec}:\Omega\times I\times\R^n\times\R^{d_{\bm{\mu}}}\to\R^{d_{\bm u}}$, is a decoder tasked with reconstructing the solution of Equation \eqref{eq:abstract} at arbitrary space–time locations and parameter instances.
Specifically, given a spatial coordinate $\bm{x}\in\Omega$, a time $t\in I$, the signal $\bm{\mu}(t)\in\R^{d_{\bm\mu}}$, and the latent state $\bm{s}(t)\in\R^n$ obtained by integrating \eqref{eq:latent_ODE_1} up to time $t$, the decoder provides the approximation
\begin{equation}\label{eq:output_ldnet}
    \bm{u}_h(\bm{x},t;\bm\mu)\approx\NN{dec}\bigl(\bm x, t,\bm{s}(t), \bm\mu(t)\bigr)\in\R^{d_{\bm u}}.
\end{equation}
Notice that the latent dimension $n$ is a hyperparameter that can be chosen using some \emph{a priori} information on the reduced dimensionality, (e.g., $N_{\bm\mu}+1$ for a time-dependent problem with constant signals \cite{fresca_comprehensive_2021}).

\glspl{ldnet} avoid the need for compressing high-dimensional solution fields into low-dimensional latent variables, leading to reduced computational overhead.
As meshless architectures, they leverage weight sharing across query points, enabling evaluation at arbitrary spatio-temporal coordinates.
Furthermore, in contrast to many conventional \gls{mor} methodologies, \glspl{ldnet} provide native support for time-dependent signals.

Despite these advantages, two primary limitations persist: (i) the capability to handle problems with varying initial conditions or input fields, and (ii) the lack of geometric information from the full-order space, which arises directly from their meshless formulation.

To address the latter issue, the following subsection introduces \glspl{ldgcn}, a graph-based extension of the \gls{ldnet} framework that preserves geometric inductive biases.

\subsection{Latent Dynamics Graph Convolutional Networks}\label{subsec:ldgcn}
\begin{figure}[htbp]
    \centering
    \includegraphics[width=0.8\textwidth]{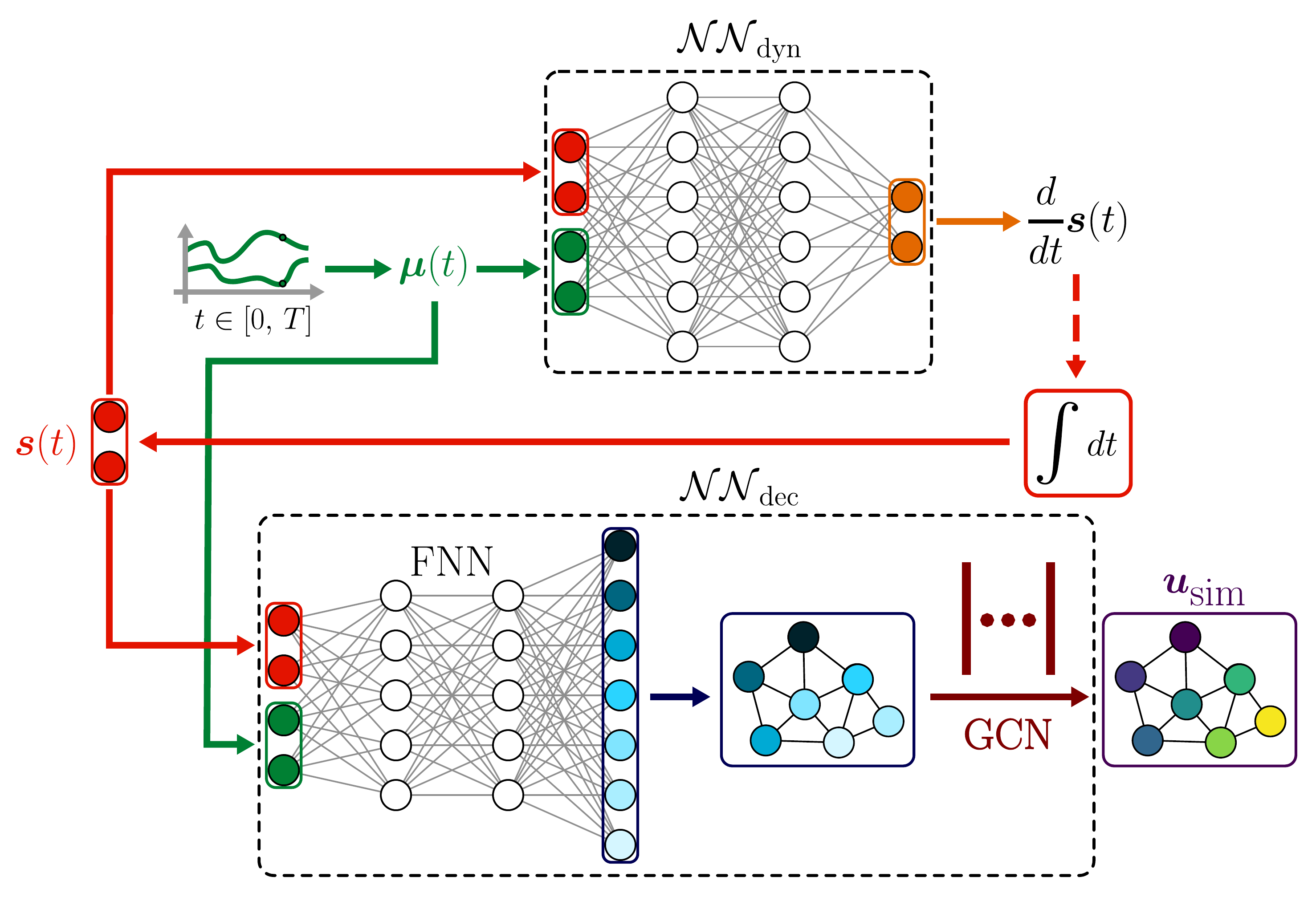}
    \caption{Schematic representation of the \gls{ldgcn} architecture.}
    \label{fig:architecture}
\end{figure}

Building upon the \gls{ldnet} framework introduced above, we now present the proposed \gls{ldgcn} architecture, which extends \glspl{ldnet} incorporating a graph-based decoder to deal with parametric time-dependent \glspl{pde} defined on complex domains with unstructured meshes.
As in \gls{ldnet}, the model adopts a two-branch structure in which the dynamics of the system is evolved in a low-dimensional latent space through a \gls{node}, while a second network reconstructs the high-dimensional solution.
In the following, we describe the two components of the \gls{ldgcn} architecture, for which visual representation is provided in Figure \ref{fig:architecture}, highlighting both the similarities with \gls{ldnet} and the advantages introduced by the graph-based decoder.
As in \gls{ldnet}, the first network $\NN{dyn}:I\times\R^n\times\R^{d_{\bm\mu}}\to\R^n$ acts as a \gls{node} and models the latent dynamics of the system according to Equation \eqref{eq:latent_ODE_1}.
The distinctive feature of \gls{ldgcn} lies in the decoding stage, where the meshless decoder of \glspl{ldnet} is replaced with a graph convolutional network defined on the mesh associated with the full-order model.
More in detail, the second network $\NN{dec}:I\times\R^n\times\R^{d_{\bm\mu}}\to\R^{N_h\times d_{\bm u}}$ is designed to reconstruct the values of the solution $\bm{u}_h(t;\bm\mu)$ to \eqref{eq:abstract} at any time and signal instance on the whole computational grid.
Given $(t,\bm\mu)\in I\times\mP$, then $\NN{dec}$ takes as input the time $t$, the signal $\bm{\mu}(t)$, and the latent state $\bm{s}(t)$, dropping the space coordinate $\bm x$ since the reconstruction is global, resulting in the following approximation:
\begin{equation}\label{eq:output_ldgcn}
    \bm{u}_h(t;\bm\mu)\approx\bm{u}_\text{sim}(t;\bm\mu)=\NN{dec}\bigl(t,\bm{s}(t), \bm\mu(t)\bigr)\in\R^{N_h\times d_{\bm u}}.
\end{equation}
We choose $\NN{dec}$ as a graph decoder inspired by that of \gls{gca} \cite{pichi_graph_2024}, to fully take advantage of the geometric information encapsulated in the grids associated to the full-order solutions.
Given the latent state $\bm{s}(t)$, the decoder uses a fixed number of fully-connected, feedforward layers to connect the bottleneck to the graph, and the decoded values of the latent states on the graph are then used to perform spatial convolutions as in Subsection \ref{subsec:GNN}.

Summarizing, given a time-dependent signal $\bm\mu\in\mP$, \gls{ldgcn} reconstructs $\{\bm u_h(t;\bm\mu)\}_{t\in I}$ by means of the following \gls{ode} system:
\begin{equation}\label{eq:latent_ODE_2}
    \begin{cases}
         \frac{d}{dt}\bm{s}(t)=\NN{dyn}\bigl(t,\bm{s}(t),\bm{\mu}(t)\bigr),& t\in(0,T]\\
         \bm{s}(0)=\bm0,\\
         \bm{u}_\text{sim}(t;\bm\mu)=\NN{dec}\bigl(t,\bm{s}(t),\bm\mu(t)\bigr),&t\in [0,T].
    \end{cases}
\end{equation}

The discretization of Equation \eqref{eq:latent_ODE_2} is carried out using numerical integration schemes for \glspl{ode}.
In this work, we employ the explicit Euler method, although more sophisticated approaches could be used, e.g., other Runge--Kutta methods \cite{quarteroni_numerical_2007}.
The integration step size $\Delta t$ does not necessarily coincide with the sampling interval of the high-fidelity data; in such cases, the values of the (discrete) signals $\bm\mu(t)$ at time instances $t\notin I_{h,\text{train}}$ can be obtained via in-time interpolation.

The network $\NN{dyn}$ exhibits a recursive nature: once the dynamics has been integrated for one time unit, say, from $t_n$ to $t_{n+1}$ according to \eqref{eq:latent_ODE_1}, the resulting latent state $\bm s(t_{n+1})$ can be fed back into the network to advance the solution further in time.
This property distinguishes \gls{ldnet} and \gls{ldgcn} from architectures such as \gls{gca}, as it enables causality and dealing with time-dependent signals.
Indeed, two distinct signals that coincide at a given time instant $t$ may still produce two different latent states at $t$.
The recursive formulation accounts for the implicit dependence of the latent state $\bm{s}(t)$ on the entire past trajectory $\{\bm\mu(\tau)\}_{\tau \le t}$.
To emphasize this dependence, we will often write the latent state as $\bm{s}(t;\bm\mu)$.

Unlike meshless methodologies, \gls{ldgcn} operates directly on the computational mesh associated with the full-order model.
On one side, this choice enables the architecture to explicitly leverage the geometric and topological information encoded in the grid, thereby improving its ability to handle complex domains, unstructured meshes, and geometry-dependent dynamics.

On the other side, as reflected by output dimensions in Equations \eqref{eq:output_ldnet} and \eqref{eq:output_ldgcn}, the drawback of this approach is that the grids associated to \gls{ldgcn} might be high-dimensional.
In particular, since the fully-connected layers of the decoder produce outputs in $\R^{N_h\times d_{\bm u}}$, the number of trainable parameters increases linearly with the number of grid points, scaling as $O(N_h)$, unlike the case of \gls{ldnet}.
The larger number parameters, which is independent of the physics of the problem, can improve robustness and reduce reliance on hyperparameter tuning, which is instead extensively employed in the original \glspl{ldnet} work.
However, it also makes the approach impractical for problems with a very large number of degrees of freedom.
Additionally, the \gls{ldgcn} decoder relies on the assumption made in Subsection \ref{subsec:problem_statement} of fixed number of grid points and mesh connectivity.
While this assumption remains compatible with geometrically parameterized meshes, it represents a limitation that we intend to address in future works.
Possible approaches to mitigating this issue include the use of a multifidelity decoder, such as that proposed in \gls{gfn} \cite{morrison_gfn_2024}, as well as the incorporation of graph pooling and unpooling operations \cite{bronstein_geometric_2021}.

In general, the choice between \gls{ldnet} and \gls{ldgcn} should be guided by the problem at hand: problems on simple geometries can be well-handed by \glspl{ldnet}, whereas cases with complex geometric parameterization are likely going to benefit from topological information from the mesh, hence suggesting the use of \gls{ldgcn}.

\begin{remark}\label{rmk:signature}
To recover simple, potentially autonomous systems, it is possible to omit time as an input to $\NN{dyn}$.
Likewise, the decoder does not necessarily need to depend on $t$ or $\bm\mu(t)$. 
Accordingly, for visualization purposes, we dropped the time dependency for both $\NN{dyn}$ and $\NN{dec}$ in Figure~\ref{fig:architecture}.
The decision to include or exclude such dependencies can be guided by prior physical knowledge of the system or determined through hyperparameter tuning \cite{regazzoni_learning_2024}.
\end{remark}

To train the architecture, a suitable loss function needs to be defined.
Denoting the weights of $\NN{dec}$ and $\NN{dyn}$ with $\bm w_\text{dec}$ and $\bm w_\text{dyn}$ respectively, we define the loss metric to be optimized as
\begin{equation}\label{eq:loss}
    \mathcal{L}(\bm u_h, \bm u_\text{sim}; \bm w_\text{dec}, \bm w_\text{dyn})=\mathcal{L}_\text{err}(\bm u_h, \bm u_\text{sim})+\lambda\mathcal{L}_\text{reg}(\bm w_\text{dec}, \bm w_\text{dyn}),
\end{equation}
where $\mathcal{L}_\text{err}: \R^{N_h\times d_{\bm u}}\times \R^{N_h\times d_{\bm u}}\to\R^+$ denotes some dissimilarity measure between the real and simulated solutions, while $\mathcal{L}_{reg}$ is the $L^1$-regularization of the weights of the network, with $\lambda\geq0$ a hyperparameter.
The first component of the loss ensures a faithful reconstruction of the high-fidelity solutions, while the second is often used in deep learning to prevent overfitting \cite{prince_understanding_2023}.
The training procedure is summarized in Algorithm \ref{alg:ld_gcn}.

\begin{algorithm}[ht]
\caption{Training algorithm for \gls{ldgcn}}
\label{alg:ld_gcn}
\SetAlgoLined
\KwIn{$\mathcal{D}_\text{train}=\{(t, \bm\mu),\,\bm{u}_h(t;\bm\mu)\}_{(t,\bm\mu)\in\mathcal{T}_\text{train}}$ training dataset, $\Delta t$ step size}
\KwOut{$\bm w_\text{dyn}, \bm{w}_\text{dec}$ trained weights of $\NN{dyn}$ and $\NN{dec}$}
\For{epoch = 1, 2, \dots}{\label{line:epoch}
    $\mL_{\text{tot}} \gets 0$\;\label{line:loss_zero}
    \For{$i =0$ \KwTo $N_{\bm\mu,\mathrm{train}}-1$\tcp*{Perform time stepping for each simulation}}{
        $\bs s(0) \gets \bs0$\;
        $\bm u_{\text{sim}}
        \gets \NN{dec}(t_0, \bm s(0), \bm\mu_i(t_0))$\tcp*{Decode initial condition}
        
        $\mL_{\text{tot}}\gets \mL_{\text{tot}}+\mL(\bm u_h(t_0;\bm\mu_i), \bm u_{\text{sim}})$\tcp*{Initial condition loss}\label{line:loss_IC}
        \For{$k = 0$ \KwTo $N_{t,\mathrm{train}}-1$\tcp*{For loop on snapshots}}{
            $h \gets t_{k+1}-t_k$\tcp*{Time interval between snapshots}
             $N_\text{int}\gets \lfloor\frac{h}{\Delta t}\rfloor$\tcp*{Number of integration steps}
            \For{$j = 0$ \KwTo $N_{\mathrm{int}} - 1$\tcp*{Latent time integration}}{
                $t_j \gets t_k + j \Delta t$\;
                $\bm\mu_j\gets\mathtt{interpolate}(j,\bm\mu_i(t_k),\bm\mu_i{(t_{k+1})})$\tcp*{Interpolate signal if needed}
                $\dot{\bm s}(t_j) \gets \NN{dyn}(t_j,\bm s(t_j), \bm \mu_j)$\;
                $ \bm s(t_{j+1}) \gets \bm s(t_j) + \Delta t \cdot \dot{\bm s}(t_{j})$\tcp*{Forward Euler step}}
           
            $\bs{u}_\text{sim} \gets \NN{dec}(t_{k+1}, \bs s(t_{k+1}), \bm\mu(t_{k+1}))$\tcp*{Decode solution}
            $\mL_\text{tot}\gets \mL_\text{tot} + \mL(\bs u_h(t_{k+1};\bm\mu_i), \bs{u}_\text{sim})$\tcp*{Loss update}
        }
    }
    $\mathtt{optimizer\_step}(\bs{w},\, \nabla_{\bs w}\mL_\text{tot})$\tcp*{Weights update}
}
\end{algorithm}

The online phase follows the same sequence of operations described in the lines $8-19$ of the algorithm.
Training-specific steps, namely iteration over epochs, evaluation of the loss function, and optimizer updates, are omitted at inference time.
During evaluation, the model does not require access to input–output pairs as in $\mathcal{D}_\text{train}$; instead, it is queried only at the desired time–parameter instances, collected in the set $\mathcal{T}_\text{test}$.

\begin{remark}[Scaling]\label{rmk:scaling}
To streamline the training procedure, we scale the snapshot matrices before passing them to the network.
For each node $i\in\{1,\dots,N_h\}$, we define the normalization constants
\begin{align*}
\alpha_{1,i} &= \frac{1}{2} \Big( \max_{(t,\bm\mu)\in\mathcal{T}_\text{train}} u^i_h({t;\bm\mu})\, + \min_{(t,\bm\mu)\in\mathcal{T}_\text{train}} u^i_h({t;\bm\mu})\Bigr), \\
\alpha_{2,i} &= \frac{1}{2} \Big( \max_{(t,\bm\mu)\in\mathcal{T}_\text{train}} u^i_h({t;\bm\mu})\, - \min_{(t,\bm\mu)\in\mathcal{T}_\text{train}} u^i_h({t;\bm\mu})\Bigr), 
\end{align*}
where $u^{i}_h(t;\bm\mu)$ denotes the value of $\bm u$ at time $t$ corresponding to the signal $\bm\mu$ at the $i$-th mesh element, with $(t,\bm\mu)\in I_h\times\mP_h$.

The normalized solutions $\tilde{\bm u}_h(t;\bm\mu)$ (or each of their components for vector valued solutions) and their inverse scaling are then respectively given for all $(t,\bm\mu) \in I\times\mP$ by\footnote{We perform a similar scaling to the one exploited in \cite{regazzoni_learning_2024}, but do not scale the input signals and times.}
\begin{equation*}
\tilde{ u}_h^i(t;\bm\mu) = \frac{u^i_h(t;\bm\mu) - \alpha_{1,i}}{\alpha_{2,i}}, \quad \text{and} \quad
u^i_h(t;\bm\mu) = \tilde{u}^i_h(t;\bm\mu) \, \alpha_{2,i} + \alpha_{1,i}, \quad  \forall i=1,\dots,N_h.
\end{equation*}
\end{remark}

\subsubsection{Theoretical guarantees}\label{subsubsec:theorems}
Since the early developments of machine learning, a major research direction has focused on establishing the existence of model architectures that can approximate a target mapping with arbitrarily small error \cite{tikk_survey_2003}.
These results, generally known as \glspl{uat}, provide the theoretical justification for the expressive power of \glspl{nn}.
More recently, a complementary line of work has aimed at deriving error bounds for deep learning–based surrogate models \cite{hillebrecht_certified_2022,biswas_error_2022, franco_practical_2025, franco_approximation_2023}, thereby improving their reliability in scientific computing settings.

For example, in \cite{farenga_latent_2025} the authors prove the learnability of an \gls{ae}-based model whose latent dynamics is governed by a \gls{node}.
Building on such result, the goal of this section is to establish a \gls{uat} for an analogous architecture in an encoder-free setting.
To simplify the exposition, we consider the case in which the signals in $\mP$ do not depend on time, allowing us to identify $\mP$ as a subset of $\R^{d_{\bm{\mu}}}$.
Additionally, we assume that $d_{\bm u}=1$ and adopt the following assumption, as in \cite{brivio_error_2024,farenga_latent_2025, franco_deep_2022}.
\begin{definition}
A \emph{perfect embedding assumption} holds if there exists $n^\ast\ll N_h$ such that, for all $n\geq n^\ast$, there exist:
\begin{enumerate}[label=(\roman*)]
\item a decoder function $\Phi\in C^1(\R^n; \R^{N_h})$,
\item a Lipschitz-continuous function $\bm f_n:(0,T]\times\R^n\times\mP\to\R^n$, with respect to $t$ and $\bm s$ uniformly in $t$, defining the latent dynamics,
\end{enumerate}
such that the latent state $\bm s(t;\bm\mu)$ satisfies $\bm u_h(t;\bm\mu) = \Phi\left(\bm s(t;\bm\mu)\right)$ and $\dot{\bm s}(t, \bm\mu)=\bm f_n\left(t,\bm s(t; \bm \mu), \bm\mu\right)$, for all $(t,\bm\mu)\in (0,T]\times\mP$.
Additionally, for each $n\geq n^\ast$, there exists an encoder $\Psi\in C^1(\R^{N_h}; \R^n)$ such that $\bm s(t_0;\bm\mu)=\Psi\left(\bm u_h(t_0;\bm\mu)\right)$, holding for each $\bm\mu\in\mP$.
\end{definition}

\begin{corollary}[UAT for encoder-free architectures]\label{theo:uat_ldnet}
Let $I_h$ be a discretization of $I$ with step size $\Delta t$.
Assume that the initial condition of \eqref{eq:abstract} is fixed and equal to $\bm u_h(t_0)$ for all $\bm\mu\in\mP$.
If the perfect embedding assumption holds, then for each $\varepsilon >0$, there exist:
\begin{enumerate}[label=(\roman*)]
    \item a dynamics network $\mathcal{NN}_\mathrm{dyn}: I\times\R^{d_{\bm\mu}}\to\R^n$ with at most $O(n\varepsilon^{-(d_{\bm\mu}+1)})$ active weights,
    \item a graph decoder $\mathcal{NN}_\mathrm{dec}:\R^n\to\R^{N_h}$, with at most $O(N_h\varepsilon^{-n})$ active weights,
    \item $\Delta t^\ast$ depending on $\mathcal{NN}_\mathrm{dyn}$,
\end{enumerate}
such that
\begin{equation*}
    \sup_{k\in\{1, \dots, N_{ t}\}} \|\bm u_h(t_k;\bm\mu) - \bm u_\mathrm{sim}(t_k;\bm\mu)\|\leq\varepsilon,\quad\forall\, \Delta t\leq\Delta t^\ast,\quad\forall\, \bm\mu\in\mP,
\end{equation*}
with
\begin{equation}\label{eq:corollary}
    \bm u_\mathrm{sim}(t_k;\bm\mu) = \mathcal{NN}_\mathrm{dec}\left(\bm s(t_k;\bm\mu)\right) = \mathcal{NN}_\mathrm{dec}\left(\Delta t \sum_{j=0}^k \omega_j\mathcal{NN}_\mathrm{dyn}(t_j, \bm\mu)\right),\quad\forall k=1,\dots,N_{ t},
\end{equation}
where $\{(\omega_j, t_j) \,|\, j=0,\dots, k \}$ are the quadrature weights and nodes.
\end{corollary}

Before proving the previous result, we briefly highlight the main steps for the proof of \cite[Theorem~1]{farenga_latent_2025}, providing the main ideas and omitting technical details to familiarize the reader with the key result on which the corollary is based.
The core idea is that, under the perfect embedding assumption, one can define a latent dynamical system governed by $\bm f_n$ such that (i) encoding the initial condition, (ii) evolving the latent dynamics, and (iii) decoding the field is equivalent to evolving the full-order system directly.
Suitable \glspl{uat} \cite{guhring_approximation_2021} are then invoked to provide the existence of \gls{nn} approximations of the encoder $\Psi$, the decoder $\Phi$, and the latent vector field $\bm f_n$.
The final step of the proof consists in deriving an error bound when the learned latent dynamics is integrated in time.

To prove Corollary~\ref{theo:uat_ldnet}, we rely on the above result.
What remains to be shown is that the construction also applies to encoder-free architectures, and that the decoder can be realized as a \gls{gnn} rather than as a fully connected network.

\begin{proof}
Under the assumptions of the corollary, the existence of fully connected networks ${\mathcal{NN}}_\text{dec}$ and $\NN{dyn}$ 
follows directly from \cite[Theorem~1]{farenga_latent_2025}.
Since we may assume without loss of generality that $\Psi\left(\bm u_{h}(t_0)\right)=\bm 0_{\R^n}$, the same theorem also guarantees the validity of \eqref{eq:corollary} due to the fixed initial condition.

Moreover, when the initial state is fixed, the encoder $\Psi$ does not enter the latent evolution, and the resulting architecture is effectively encoder-free.
This establishes the existence of a feedforward architecture whose evolution is governed by \eqref{eq:latent_ODE_2}.
It remains to show that $\NN{dec}$ can be seen as a specific instance of the decoder employed in \gls{gca}.

Specifically, the graph decoder in \gls{gca} consists of a fully-connected network followed by a sequence of message-passing layers similar to those described in Equation \eqref{eq:gcn} (more details can be found in Appendix \ref{sec:appendix}).
Due to the presence of skip-connections \cite{prince_understanding_2023}, the output of each layer can be written as $\bm h^{(k-1)}_u+\bm h^{(k)}_u$.
If all the convolutional weights and biases in \eqref{eq:gcn} are set to $0$ and the activation $\sigma$ is chosen as the identity, then $\bm h^{(k-1)}_u+\bm h^{(k)}_u=\bm h^{(k-1)}_u$, implying that each convolutional layer acts as the identity map.
As a result, $\NN{dec}$ can be regarded as a graph decoder in which all the parameters of the (arbitrary) number of convolutional layers are identically set to $0$.\qedhere
\end{proof}

Although Corollary \ref{theo:uat_ldnet} assumes the existence of an encoder $\Psi$, \gls{ldgcn} does not introduce a learnable approximation $\Psi_\theta$ as a result of the independence of the initial condition on $\bm\mu$.

Regarding the dynamics network $\mathcal{NN}_{\text{dyn}}$ appearing in Corollary \ref{theo:uat_ldnet}, we observe that it depends only on the query point $(t, \bm\mu(t))$ and not on the latent state $\bm s(t;\bm\mu)$.
In contrast, the \gls{ldgcn} architecture also provides the latent state as an input to $\mathcal{NN}_{\text{dyn}}$.
Our design choice not only increases the expressivity of the model, but also enables the treatment of time-dependent input signals.
Indeed, this discrepancy arises from the assumption of constant signals adopted in the corollary.
As discussed previously, if this assumption were relaxed, two distinct time-dependent signals attaining the same value at a given time $t$ could nevertheless correspond to different latent states.
Since $\mathcal{NN}_{\text{dyn}}$ has access only to the pointwise value of the input signal and not to its full temporal history, incorporating $\bm s(t;\bm\mu)$ as an additional input allows the latent dynamics to depend on the state reached at time $t$, rather than solely on the instantaneous value of the input.
Ultimately, even for constant signals, this choice generalizes the architecture and facilitates the extension of universal approximation results to time-dependent signals in future work.
Finally, the signature of $\NN{dec}$ is compatible with that of \gls{ldgcn} in view of Remark \ref{rmk:signature}.

\subsection{Interpolation of latent trajectories}\label{subsec:interpolation}
When studying time-dependent systems, exploiting information about the dynamics can help to regularize the latent state, providing a meaningful and more interpretable low-dimensional representation of trajectories \cite{champion_data-driven_2019, conti_reduced_2023}.
One of the defining features of \glspl{ldnet} is the simplicity of the latent spaces resulting from training, also benefiting from the absence of an \gls{ae} architecture that learns from data only exploiting statistical information \cite{regazzoni_learning_2024}.

As we will see, the well-structured nature of the latent trajectories makes it natural to wonder whether it is possible to interpolate them and perform zero-shot predictions.
That is, whether it is possible to reconstruct the whole dynamics globally without the need to integrate in time the latent space during the online phase, thereby speeding up computations for new time and parameter values.
Moreover, such an approach could be useful even in some cases in which the latent dynamic is integrated in time.
For instance, let us assume that we want to reconstruct the evolution of a solution only for times $t\geq \tilde{t}>t_0=0$, e.g., to study long-time behavior dynamics.
If interpolation is not adopted, then the use of time-stepping requires integrating from $t_0$ up to $\tilde t$, since the initial condition $\bm{s}(0; \cdot)=\bm{0}$ cannot be changed.
Instead, using interpolation would bypass integration helping the approximation when the latent dynamics is stiff or hard to evolve using explicit methods, in addition to mitigate even further the computational cost tied to the online phase.

\begin{remark}
    Notice that the aforementioned procedure could seem in general counterintuitive, since the main goal of the proposed \gls{ldgcn} is to produce a dynamics which is consistent with the arrow of time.
    Although using interpolation might not seem compatible with this reasoning, we found it meaningful to investigate whether the main cause for the good performance of latent dynamics models, such as \gls{ldnet}, lies in the regularization introduced in the latent space by dynamical information, or in the actual time integration procedure.
\end{remark}

Let us now assume that $\mP\subset\R^{d_{\bm\mu}}$ and consider a time $t$ and an input signal $\bm{\mu}$, with $(t,\bm\mu)\in I\times \mP$.
Then it is possible to interpolate the latent states resulting from network training $\{\bm{s}(\tau;\bm{\nu})\}_{(\tau,\bm\nu)\in \mathcal T_{\text{train}}}$, obtaining an interpolant $\tilde{\bm{s}}:I\times\mP\to\R^n$.
Such an interpolant can then be evaluated at $(t, \bm\mu)$, providing an estimate of the latent state $\bm s(t;\bm\mu)$ that would be obtained using integration of the outputs of $\NN{dyn}$.
In turn, this allows one to obtain an approximation of the solution $\bm{u}_h(t; \bm{\mu})$ by decoding $\tilde{\bm{s}}(t;\bm\mu)$ using $\mathcal{NN}_\text{dec}$, i.e.,
\begin{equation*}
\begin{aligned}
\bm u_\text{interp}(t, \bm{\mu})&=\mathcal{NN}_\text{dyn}\left(t, \tilde{\bm s}(t; \bm{\mu}), \bm{\mu}\right)\\
&\approx \mathcal{NN}_\text{dyn}\left(t, {\bm s}(t; \bm{\mu}), \bm{\mu}\right)\\
&=\bm u_\text{sim}(t;\bm\mu) \approx \bm u_h(t; \bm{\mu}).
\end{aligned}
\end{equation*}

More details on the choice of the interpolants are provided in the numerical experiments.
The procedure described in this subsection can be formally justified by the following result, which essentially states that, under suitable assumptions, an error bound for the latent interpolant directly translates into an error bound for the full-dimension trajectories.
\begin{proposition}\label{prop:interpolation}
Given a norm $\|\cdot\|$ on $\R^n$, we denote by $\varepsilon(t; \bm\mu)$ the error in the $\|\cdot\|$ norm of the \gls{ldgcn} architecture at time $t$ corresponding to the parameter $\bm\mu$ with respect to the full-order solution $\bm u_h(t;\bm\mu)$ for $(t,\bm\mu)\in I_h\times\mP$.
Assume that $\NN{dec}$ is Lipschitz-continuous with constant $L$ with respect to its second argument on $I\times\R^n\times \R^{d_{\bm\mu}}$.
Moreover, assume that the latent interpolant $\tilde{\bm s}^m$ depends on a discretization parameter $m$ (e.g., the number of subintervals when using spline functions), and that it is possible to provide a bound on the interpolation error:
\begin{equation}\label{eq:latent_interpolation_error}
    \max_{(t, \bm\mu) \in I_h \times \mP_h} \left\| \bm{s}(t; \bm\mu) - \tilde{\bm{s}}^m(t; \bm\mu) \right\| \leq \delta(m),
\end{equation}
where $\delta(m)$ depends only on $m$, and not on $t$ or $\bm\mu$.

Then it follows that for all $(t, \bm\mu)\in I\times\mP$,
\begin{equation*}
\| \bm{u}_h(t;\bm\mu) - \bm{u}_\mathrm{interp}(t,\bm\mu) \| \leq L\delta(m) + \varepsilon(t; \bm\mu).
\end{equation*}
In particular, if $\delta(m) \overset{m \to \infty}{\longrightarrow} 0$, then
\begin{equation*}
\lim_{m \to +\infty}\| \bm{u}_h(t;\bm\mu) - \bm{u}_\mathrm{interp}(t;\bm\mu) \| \leq \varepsilon(t; \bm\mu).
\end{equation*}
\end{proposition}
\proof
We can express the difference between the full-order and the interpolated solutions as the sum of two different components: 
 \begin{equation}\label{eq:error_interp}
        \bm u_h-\bm u_\text{interp}=\underbrace{\bm u_h-\bm u_\text{sim}}_{\text{error using LD-GCN}}+\underbrace{\bm u_\text{sim} - \bm u_\text{interp}}_{\text{interpolation error}}.
    \end{equation}
Then, by the triangular inequality, Lipschitz condition, and Equation \eqref{eq:latent_interpolation_error}, we can write 
\begin{align*}
    \| \bm{u}_h(t;\bm\mu) - \bm{u}_\text{interp}(t; \bm\mu) \|
    &\leq \| \bm{u}_h(t; \bm\mu)  - \bm{u}_\text{sim}(t; \bm\mu)  \| + \| \bm{u}_\text{sim}(t; \bm\mu)  - \bm{u}_\text{interp}(t; \bm\mu)  \| \\
    &\leq \varepsilon(t; \bm\mu) + \| \NN{dec}\left(t,\bm{s}(t; \bm\mu), \bm\mu\right) - \NN{dec}\left(t,\tilde{\bm{s}}^m(t; \bm\mu),\bm\mu\right) \| \\
    &\leq \varepsilon(t; \bm\mu) + L \| \bm{s}(t; \bm\mu)  - \tilde{\bm{s}}^m(t; \bm\mu)  \| \\
    &\leq \varepsilon(t; \bm\mu) + L\delta(m). \qedhere
\end{align*}

\begin{remark}
Notice that $\NN{dyn}$ can be queried for each point in the (non-discretized) domain $I \times \mP$.  
Then it is possible to further refine the grid used for interpolation at a latent level by simulating additional latent trajectories $\bm{s}(t; \bm\mu)$, for the desired $(t, \bm\mu) \in I \times \mathbb{P}$, e.g., by considering a finer discretization of $I$.  
Therefore, if the result of Proposition \ref{eq:latent_interpolation_error} holds for all new values of time and parameters that are considered and the interpolation is sufficiently accurate, then the surrogate approximation exploiting latent interpolation has a precision similar to the original \gls{ldgcn} architecture, i.e., $\|\bm u_h(t; \bm\mu) - \bm u_\text{sim}(t; \bm\mu)\|\approx \|\bm  u_h(t; \bm\mu) - \bm{u}_\text{interp}(t; \bm\mu)\|$.
\end{remark}

\begin{remark}[Extrapolation in time]\label{rmk:interp_extrap}
Performing interpolation might raise issues if $t>\max_{\tau\in I_{h,\text{train}}}\tau$, as it would be required to extrapolate using $\tilde{\bm{s}}$.
However, some interpolants, such as spline functions, are not defined outside of the convex hull of the interpolation points.
The problem can be solved by considering one of the following two strategies: 
\begin{enumerate}[label=(\roman*)]
    \item\label{item:integration} integrate-then-interpolate: integrate the training trajectories up to $t$ for $\bm\mu\in\mP_{h,\text{train}}$ and then interpolate to allow evaluation of $\bm s(t;\bm\mu)$ for $\bm\mu\notin\mP_{h,\text{train}}$;
    \item\label{item:extrapolation} interpolate-then-extrapolate: choose an interpolator $\tilde{\bm{s}}$ that is capable of performing extrapolation in time, e.g.,\ \gls{gpr} \cite{rasmussen_gaussian_2005}.
\end{enumerate}
\end{remark}

As discussed in Subsection \ref{subsec:ldgcn}, when considering time-dependent signals, any mapping to $\bm{s}(t;\bm\mu)$ defined only in terms of the pair $(t,\bm{\mu}(t))$ becomes ill-posed: distinct input signals may coincide at time $t$ while differing at earlier times, resulting in different latent states.
Since the interpolation strategies employed in this study rely on pointwise mappings $(t,\bm\mu(t))\mapsto\bm s(t;\bm\mu)$, we restrict this analysis to signals that are constant in time.
Under this assumption, the latent state is uniquely determined by the pair $(t,\bm{\mu})$, ensuring that the interpolation procedure is well-defined.
\section{Numerical results}\label{sec:results}
In this section, we examine the performance of \gls{ldgcn} on different benchmarks of increasing complexities governed by PDEs with both physical and geometric parameters.
We investigate the results to highlight the ability of \gls{ldgcn} to accurately reconstruct full-order solutions and to extrapolate in time for parameter values not seen during training.
We also examine the behavior of the latent trajectories to assess their regularity, perform zero-shot interpolation as described in Subsection \ref{subsec:interpolation}, and evaluate the interpretability of latent trajectories to even discover complex bifurcating behavior linked to the loss of uniqueness for the solution.
This way, we aim at giving a comprehensive perspective on the performance of the newly introduced architecture w.r.t.\ the state-of-the-art and related methodologies, while trying to overcome several current challenges and providing a more suitable and consistent framework to deal with general time-dependent and parameterized PDEs.

The implementation of the \gls{ldgcn} architecture is based on \cite{pichi_graph_2024} and relies on PyTorch Geometric~\cite{fey_pyg_2025}.
Data collection was performed with the \gls{fe} method \cite{quarteroni_numerical_2017} implemented using FEniCS \cite{logg_automated_2012} and RBniCS \cite{rozza_real_2024}.
The parameters of all the architectures considered are summarized in Table \ref{tab:network_parameters} in Appendix \ref{sec:appendix}.

Following the notation adopted in Section \ref{sec:methodology}, we introduce the error metrics described in the numerical experiments.
Given the full-order and reduced-order solutions $\bs{u}_h, \bs{u}_\text{sim}:I_h\times\mathbb{P}_h\to\mathbb{R}^{N_h\times d_{\bs{u}}}$, for each $(t,\bs{\mu})\in I_h\times\mathbb{P}_h$ we define the relative error field $\bs{e}_\text{rel}:I_h\times\mathbb{P}_h\to\mathbb{R}^{N_h\times d_{\bs{u}}}$ and its norm $\varepsilon_\text{rel}$ as 
\begin{equation}\label{eq:rel_ell}
    \bs{e}_\text{rel}(t; \bs{\mu})= \frac{\bs{u}_h(t; \bs{\mu})-\bs{u}_\text{sim}(t; \bs{\mu})}{\|\bs{u}_h(t; \bs{\mu})\|_{2}}, \qquad \text{and} \quad
    \varepsilon_\text{rel}(t; \bs{\mu})=\|\bs{e}_\text{rel}\|_{2},\quad\forall \,(t,\bs{\mu})\in I_h\times\mathbb{P}_h.
\end{equation}
We also define the maximum and mean relative errors across the entire dataset as
\begin{equation*}
\varepsilon_\text{max}=\max_{(t,\bs{\mu})\in I_h\times\mathbb{P}_h}\varepsilon_\text{rel}(t;\bs{\mu}), \qquad \varepsilon_\text{mean}=\frac{\sum_{(t,\bs{\mu})\in I_h\times\mathbb{P}_h}\varepsilon_\text{rel}(t;\bs{\mu})}{|I_h\times\mathbb{P}_h|}.
\end{equation*}

\subsection{Advection-diffusion equation}\label{subsec:adv_diff}
We start by considering two benchmarks for a parameterized advection-diffusion equation, the first one involving physical parameters and the second one geometric parameters \cite{franco_deep_2023, chen_time_2025}.

\subsubsection{Problem definition and data collection}
Consider a parameter space $\mathbb{P}\subset\R^2$ and a collection of parameterized domains $\{\Omega_{\bs{\mu}}\}_{\bs{\mu}\in\mP}$, with $\Omega_{\bs{\mu}}\subset\R^2$ for each $\bs\mu\in\mP$.
Let $\kappa\geq0$, $f:\Omega_{\bm\mu}\times(0,T]\times\mP\to\R$, and $\bs{\beta}:[0,T]\times\mP\to\R^2$.
Given $\bs\mu\in\mP$, the problem considered consists in finding $u: \Omega_{\bm\mu} \times [0, T] \to \mathbb{R}$ such that:
\begin{equation}\label{eq:SquareAdvStrong1}
\begin{cases}
\displaystyle\frac{\partial u}{\partial t}(\bs{x}, t)
- \nabla \cdot (\kappa \nabla u)(\bs{x}, t)
+ \bs{\beta}(t; \bs{\mu}) \cdot \nabla u(\bs{x}, t)
= f(\bs{x}, t;\bm\mu), & \forall(\bs{x},t)\in\Omega_{\bs{\mu}}\times(0,T],\\
u(\bs{x}, t) = u_D(\bs{x}), &\forall(\bs{x}, t)\in\Gamma_D \times (0, T],\\
u(\bs{x}, 0) = u_0(\bs{x}), &\forall\bs{x}\in\Omega_{\bs{\mu}},
\end{cases}
\end{equation}
where $\Gamma_D$ denotes the portion of the boundary on which Dirichlet boundary conditions are imposed, while we assume homogeneous Neumann boundary conditions to hold where no conditions are explicitly written. 

For all our experiments, in Equation \eqref{eq:SquareAdvStrong1} we set $\kappa=0.1,\, T=2$, $f \equiv 0$, and compatible boundary and initial conditions $u_D(\bs{x}) = u_0(\bs{x}) = (x_0 - 1)^2 + (x_1 - 1)^2$.

The numerical high-fidelity approximation has been obtained, for each value of $\bs\mu\in\mP_h$, via \gls{fe} method with Lagrange space $\mathcal{P}^1$.
Concerning the temporal discretization, time stepping was performed using the implicit Euler method \cite{quarteroni_numerical_2007}, with a step-size $\Delta t=2\cdot10^{-2}$, resulting in $I_h = \{k\cdot \Delta t,\,\text{for}\, 0\leq k\leq \frac{T}{\Delta t}\}$.

In both test cases considered, denoted with \gls{sa} and \gls{mh}, we trained a \gls{ldgcn} model and a \gls{gca} model to compare their performance on this problem.
As explained in Section \ref{sec:methodology}, while \gls{ldgcn} allows to use $\bm{\beta}(t;\bm{\mu})$ in \eqref{eq:SquareAdvStrong1} directly as input signal, this is not possible with \gls{gca}.
To ensure a fair comparison, therefore, we always provide $(t,\bm{\mu})$ as input to both architectures.

\subsubsection{Square advection (\gls{sa})}\label{subsubsec:SA}
The first test case that we tackle involves only physical parameters.
In particular, we set the parameterized advection term in \eqref{eq:SquareAdvStrong1} as $\bs{\beta}(t, \bs{\mu}) = [\mu_1 (1 - t), \mu_2 (1 - t)]$, and keep the domain fixed for all $\bs\mu\in\mP$ as $\Omega_{\bs\mu}\equiv\Omega=(0,1)^2$.
The parametric space was defined as $\mP=[-1,1]^2$ and the simulations were performed considering a discretization $\mP_h$ of $\mP$ consisting of a grid with $5$ equispaced values for both parametric $\mu_1$ and $\mu_2$, for a total of $N_{\bm\mu}=25$ simulations.
The domain was discretized using a mesh of $1472$ nodes.

To train the \gls{ldgcn} and \gls{gca} models, we randomly selected $N_{\bm\mu,\text{train}}=18$ trajectories corresponding to $75\%$ of the dataset\footnote{The same sampling has been used for the two architectures to obtain consistent and comparable results.}.
Moreover, the last $25\%$ of each dynamics was excluded from the training to be able to assess the in-time extrapolation capabilities of each network\footnote{The first snapshot has been removed from each trajectory for consistency with the other test cases considered.}, so that the last time instance seen by the network corresponded to $T_\text{train}=1.5$.
Notice that this procedure resulted in using approximately half of the original dataset: the networks only saw $1350$ of the original $2500$ snapshots during training.

\begin{figure}[htbp]
    \centering
    \begin{subfigure}[c]{0.53\textwidth}
        \centering
        \includegraphics[width=\textwidth]{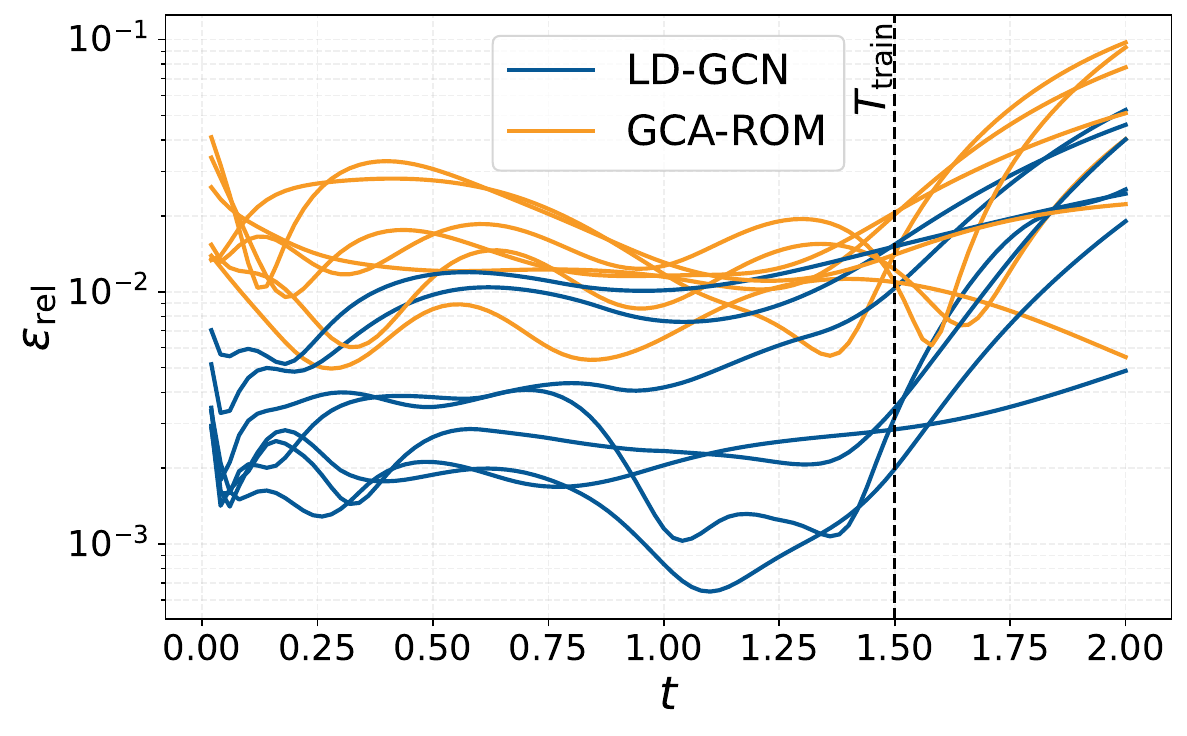}
        \caption{Relative error for \gls{ldgcn} and \gls{gca}}
        \label{fig:error_gca_SA}
    \end{subfigure}
    \hfill
    \begin{subfigure}[c]{0.44\textwidth}
        \centering
        \includegraphics[width=\textwidth]{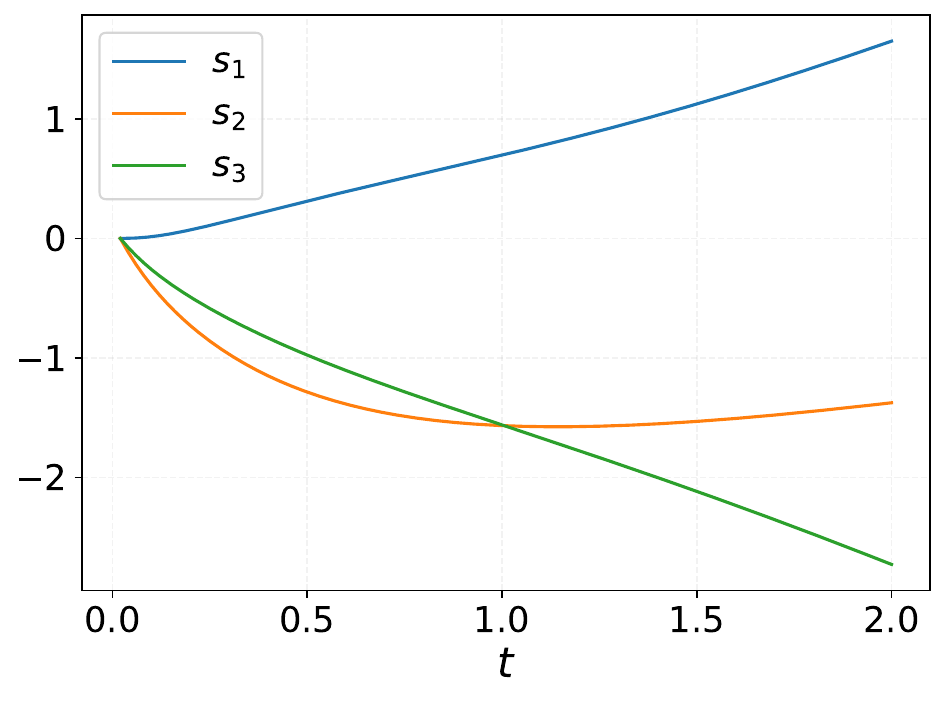}
        \caption{Latent state dynamics for $\bs{\mu}=(-1, 0.5)$}
        \label{fig:latents_SA}
    \end{subfigure}
    \caption{Relative errors for all test trajectories and latent evolution w.r.t.\ time.}
    \label{fig:SA_subfigures}
\end{figure}

We start by showing in Figure \ref{fig:error_gca_SA} the relative errors of \gls{ldgcn} and \gls{gca} architectures for trajectories belonging to the test set for $n=3$.
The plot shows great accuracy in recovering the dynamical evolution and superior performance of \gls{ldgcn} architecture over the \gls{gca} one, which exhibits an improvement of around one order of magnitude.  
Moreover, the results prove that the architecture is capable of successfully extrapolate both in time and parameters, with maximum relative error equal to $\varepsilon_{\text{max}}=5.26\cdot10^{-2}$, even for unseen trajectories beyond the final training time $T_\text{train}$.
We underline that, as a consequence of being an encoder-free architecture, the number of parameters in \gls{ldgcn} is approximately half that of \gls{gca}, which still performs fairly well as a result of the higher number of weights.
Additionally, Figure \ref{fig:latents_SA} displays the evolution of the latent state for $\bs{\mu}=(-1, 0.5)$, clearly showing their simplicity and interpretability.

To study the sensitivity of the \gls{ldgcn} architecture with respect to the latent dimension $n$, we repeated the simulations for increasing number of latent variables, the results of which are reported in Table \ref{tab:err_gca_SA}.
It is possible to see that the architecture is robust with respect to the latent dimension, as the error remains low for all considered values of $n$.
We remark that, although the number of total parameters is much lower for \gls{ldgcn} than for \gls{gca}, the latent dynamics network $\mathcal{NN}_\text{dyn}$ has more trainable parameters than the map in \gls{gca}.
The choice not to increase the number of parameters of \gls{gca} originates from the fact that the latter is more prone to overfitting.

\begin{table}[htbp]
    \caption{Comparison of \gls{ldgcn} and \gls{gca} architectures, with mean and max relative errors and number of trainable parameters for varying values of $n$. The last two columns refer to the number of weights of each subnetwork.}
    \centering
    \begin{tabular}{|c|c|c|c|c|c|c|}
        \hline
        \cellcolor{white}Test case &
        \cellcolor{white}Architecture &
        \cellcolor{white}$n$ &
        \cellcolor{white}$\varepsilon_\text{mean}$ &
        \cellcolor{white}$\varepsilon_\text{max}$ &
        \cellcolor{white}$\NN{dyn}$/map &
        \cellcolor{white}$\NN{dec}$/\gls{ae}\\ 
        \hline
        \multirow{6}{*}{\cellcolor{white}\shortstack{Square\\Advection\\(SA)}}
            & \multirow{3}{*}{\cellcolor{white}\gls{ldgcn}} 
            & \cellcolor{gray!20}$3$  & \cellcolor{gray!20}$7.86\cdot10^{-3}$ & \cellcolor{gray!20}$5.26\cdot10^{-2}$ & \cellcolor{gray!20}\num{24813} & \cellcolor{gray!20}\num{297289} \\
            &  
            & \cellcolor{gray!5}$10$ & \cellcolor{gray!5}$5.06\cdot10^{-3}$ & \cellcolor{gray!5}$4.16\cdot10^{-2}$ & \cellcolor{gray!5}\num{25520} & \cellcolor{gray!5}\num{298689} \\
            &  
            & \cellcolor{gray!20}$15$ & \cellcolor{gray!20}$5.57\cdot10^{-3}$ & \cellcolor{gray!20}$5.89\cdot10^{-2}$ & \cellcolor{gray!20}\num{26025} & \cellcolor{gray!20}\num{299689} \\
            \cline{2-7}
            & \multirow{3}{*}{\cellcolor{white}\gls{gca}} 
            & \cellcolor{gray!5}$3$  & \cellcolor{gray!5}$1.87\cdot10^{-2}$ & \cellcolor{gray!5}$9.75\cdot10^{-2}$ & \cellcolor{gray!5}\num{8003} & \cellcolor{gray!5}\num{591909} \\
            &  
            & \cellcolor{gray!20}$10$ & \cellcolor{gray!20}$1.86\cdot10^{-2}$ & \cellcolor{gray!20}$1.13\cdot10^{-1}$ & \cellcolor{gray!20}\num{8360} & \cellcolor{gray!20}\num{594716} \\
            &  
            & \cellcolor{gray!5}$15$ & \cellcolor{gray!5}$1.74\cdot10^{-2}$ & \cellcolor{gray!5}$1.11\cdot10^{-1}$ & \cellcolor{gray!5}\num{8615} & \cellcolor{gray!5}\num{596721} \\
        \hline
        \multirow{6}{*}{\cellcolor{white}\shortstack{Moving\\Hole\\(MH)}} 
            & \multirow{3}{*}{\cellcolor{white}\gls{ldgcn}} 
            & \cellcolor{gray!20}$3$  & \cellcolor{gray!20}$1.67\cdot10^{-2}$ & \cellcolor{gray!20}$1.54\cdot10^{-1}$ & \cellcolor{gray!20}\num{24813} & \cellcolor{gray!20}\num{273169} \\
            &  
            & \cellcolor{gray!5}$10$ & \cellcolor{gray!5}$9.93\cdot10^{-3}$ & \cellcolor{gray!5}$9.30\cdot10^{-2}$ & \cellcolor{gray!5}\num{25520} & \cellcolor{gray!5}\num{274569} \\
            &  
            & \cellcolor{gray!20}$15$ & \cellcolor{gray!20}$8.81\cdot10^{-3}$ & \cellcolor{gray!20}$1.02\cdot10^{-1}$ & \cellcolor{gray!20}\num{26025} & \cellcolor{gray!20}\num{275569} \\
             \cline{2-7}
            & \multirow{3}{*}{\cellcolor{white}\gls{gca}} 
            & \cellcolor{gray!5}$3$  & \cellcolor{gray!5}$4.47\cdot10^{-2}$ & \cellcolor{gray!5}$2.14\cdot10^{-1}$ & \cellcolor{gray!5}\num{8003} & \cellcolor{gray!5}\num{543789} \\
            &  
            & \cellcolor{gray!20}$10$ & \cellcolor{gray!20}$2.88\cdot10^{-2}$ & \cellcolor{gray!20}$1.55\cdot10^{-1}$ & \cellcolor{gray!20}\num{8360} & \cellcolor{gray!20}\num{546596} \\
            &  
            & \cellcolor{gray!5}$15$ & \cellcolor{gray!5}$3.54\cdot10^{-2}$ & \cellcolor{gray!5}$1.92\cdot10^{-1}$ & \cellcolor{gray!5}\num{8615} & \cellcolor{gray!5}\num{548601} \\
        \hline
    \end{tabular}
    \label{tab:err_gca_SA}
\end{table}

\begin{figure}[htbp]
    \centering
    \includegraphics[width=\textwidth]{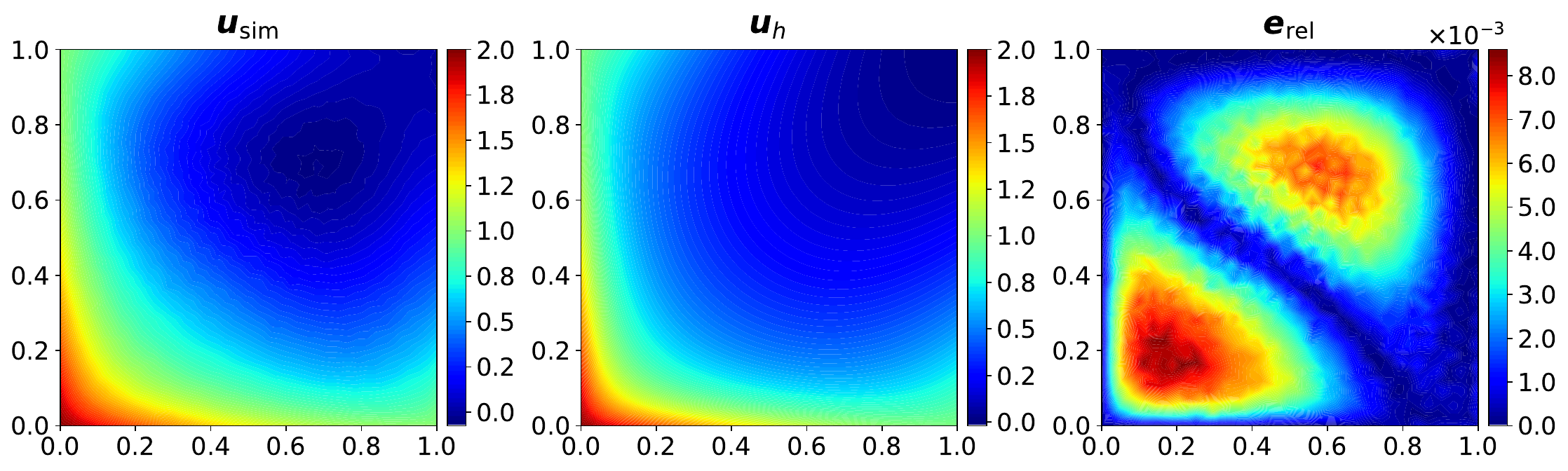}
    \caption{Plot of the simulated solution (left), the full-order one (middle) and relative $L^2$ error field (right) for $\bs{\mu}=(1, 1)$ at time $t=2$.}
    \label{fig:fields_SA}
\end{figure}

A further confirmation of the approximation properties is shown in  Figure~\ref{fig:fields_SA}, where we plot the snapshot with the highest relative error produced by \gls{ldgcn} across the entire dataset.
Indeed, its comparison with the high-fidelity solution and the corresponding relative error field $\bs{e}_\text{rel}$, demonstrates that the proposed architecture is capable of consistently evolving the features in-time well-beyond also in the extrapolation regime at final time, i.e., for $t=2$.

\subsubsection{Moving hole advection (\gls{mh})}\label{subsubsec:MH}
The second benchmark introduces the study of a geometrically parameterized domain.
In fact, the main reason for introducing a graph-based model is the flexibility to consistently exploit geometric inductive bias, naturally embedding such important information in the learning procedure of our architecture.

Consider the parametric geometry defined as $\Omega_{\bs{\mu}}=(0,1)^2\setminus S$, where $S = [\mu_1, \mu_1+0.3]\times[\mu_2, \mu_2+0.3]\subset\R^2$ represents a hole inside the square domain, and  the parametric space is given by $\bs{\mu}\in\mathbb{P}=[0.2,\,0.5]^2\subset\R^2$.
In this case, the \gls{pde} in Equation \eqref{eq:SquareAdvStrong1} only depends on geometric parameters, thus the convective term is kept fixed as $\bs{\beta}(t) = [1 - t, 1 - t]$.

The simulations were performed on a mesh consisting of $\num{1352}$ nodes, while the discretization of $\mP$ is given, as before, by a grid with $5$ equispaced values for each parametric direction, $\mu_1$ and $\mu_2$.

The number of trajectories to train the \gls{ldgcn} and \gls{gca} models, as well as the discretization of $I$ and the train/test ratios, are the same as in the \gls{sa} test case.
We remark that, as before, the first snapshot corresponding to the initial condition was removed from each simulation. Indeed, the combined presence of a boundary layer and the small norm of the solution cause the maximum relative error to be attained there even for good approximations, not allowing $\varepsilon_\text{max}$ to properly compare the performance in the extrapolation regime.

\begin{figure}[htbp]
    \centering
    \includegraphics[width=0.6\textwidth]{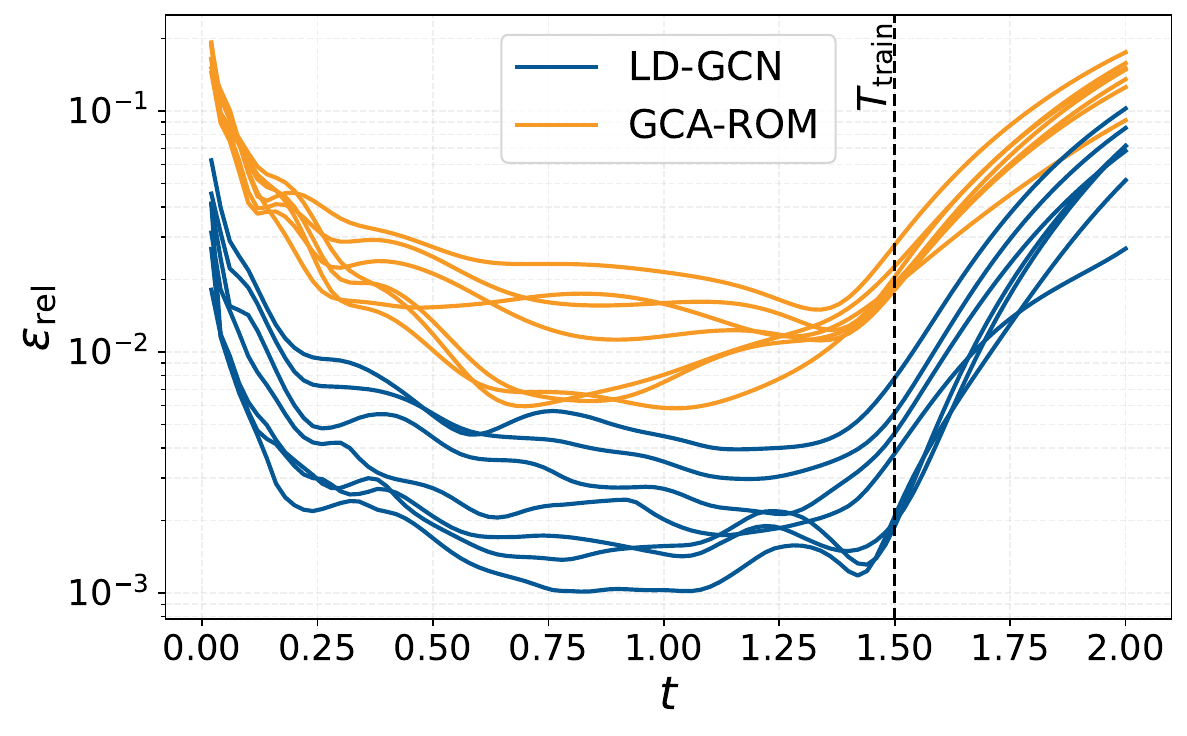}
    \caption{Relative errors for all test trajectories w.r.t.\ time.}
    \label{fig:error_gca_MH}
\end{figure}

In particular, we show in Figure \ref{fig:error_gca_MH} that, when choosing $n=15$, the relative errors on the test dataset for the \gls{ldgcn} architecture are clearly and significantly lower than for \gls{gca}, showing remarkable improvements in capturing the dynamics of the system even when dealing with complex and varying domains.
The proposed strategy is thus able to coherently treat domain complexity and dynamical behavior simultaneously, extracting and exploiting geometric and temporal features to effectively learn the latent coordinates.

Table \ref{tab:err_gca_SA} compares the relative errors of the two different architectures considered for different compression size values.
Consistently with the previous test case, the results are robust to changes in latent dimension $n$, with a slight performance improvement when increasing the expressivity of the model.

\begin{figure}[htbp]
    \centering
    
    \begin{subfigure}[b]{\textwidth}
        \centering
        \includegraphics[width=\textwidth]{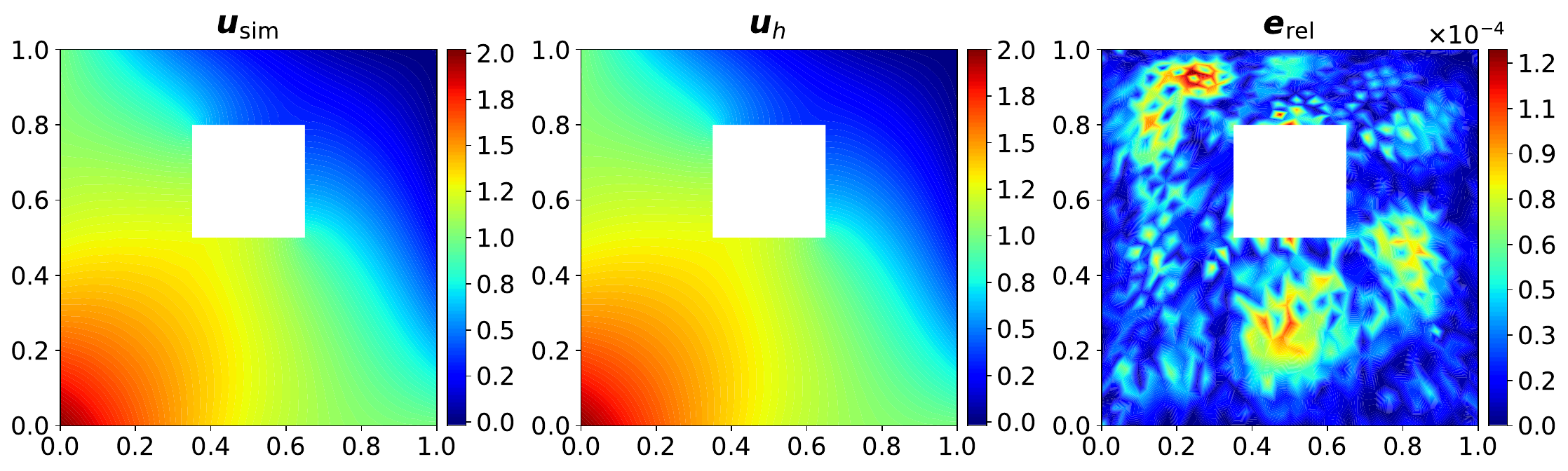}
        \caption{$\bs{\mu}=(0.35, 0.5)$ at time $t=1$}
    \end{subfigure}
    \hfill
    
    \begin{subfigure}[b]{\textwidth}
        \centering
        \includegraphics[width=\textwidth]{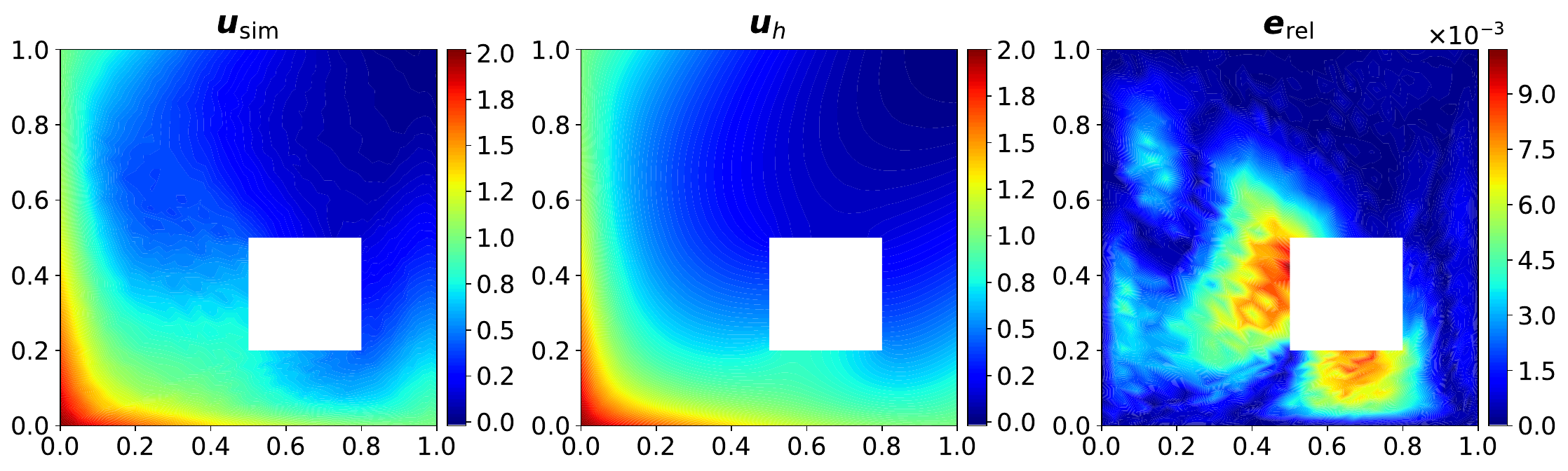}
        \caption{$\bs{\mu}=(0.5, 0.2)$ at time $t=2$}
        \label{fig:field_mh2}
    \end{subfigure}
    
    \caption{Plot of the simulated solution (left), the full-order one (middle) and relative $L^2$ error (right) for two different values of $\bs{\mu}\in\mathbb{P}_h$.} 
    \label{fig:fields_MH}
\end{figure}

Figure~\ref{fig:fields_MH} shows two instances of test snapshots defined on different configurations of the inner obstacle.
In particular, the one corresponding to the highest error produced by the \gls{ldgcn} architecture on the test dataset is reported in Figure~\ref{fig:field_mh2}, alongside the corresponding high-fidelity solution and the resulting relative error field.
Despite some inaccurate reconstruction around the moving object, where homogeneous Neumann conditions are imposed, we stress the capability of the architecture to reconstruct the overall dynamics when extrapolating for more than 30\% of the original time range.

\subsubsection{Analysis and interpolation of the latent states}\label{subsubsec:interp}
In this section, we are interested in investigating whether the strong regularity and interpretability properties of the latent trajectories obtained using \glspl{ldnet} are preserved also when exploiting an encoder-free decoder-based architecture such as \gls{ldgcn}. 
To this end, and in light of the discussion in Subsection \ref{subsec:interpolation}, we study the latent state behavior of the \gls{sa} system.

\begin{figure}[htbp]
    \centering

    \begin{subfigure}[b]{0.45\textwidth}
        \centering
        \includegraphics[width=\textwidth, trim={10cm 2cm 10cm 1cm},clip]{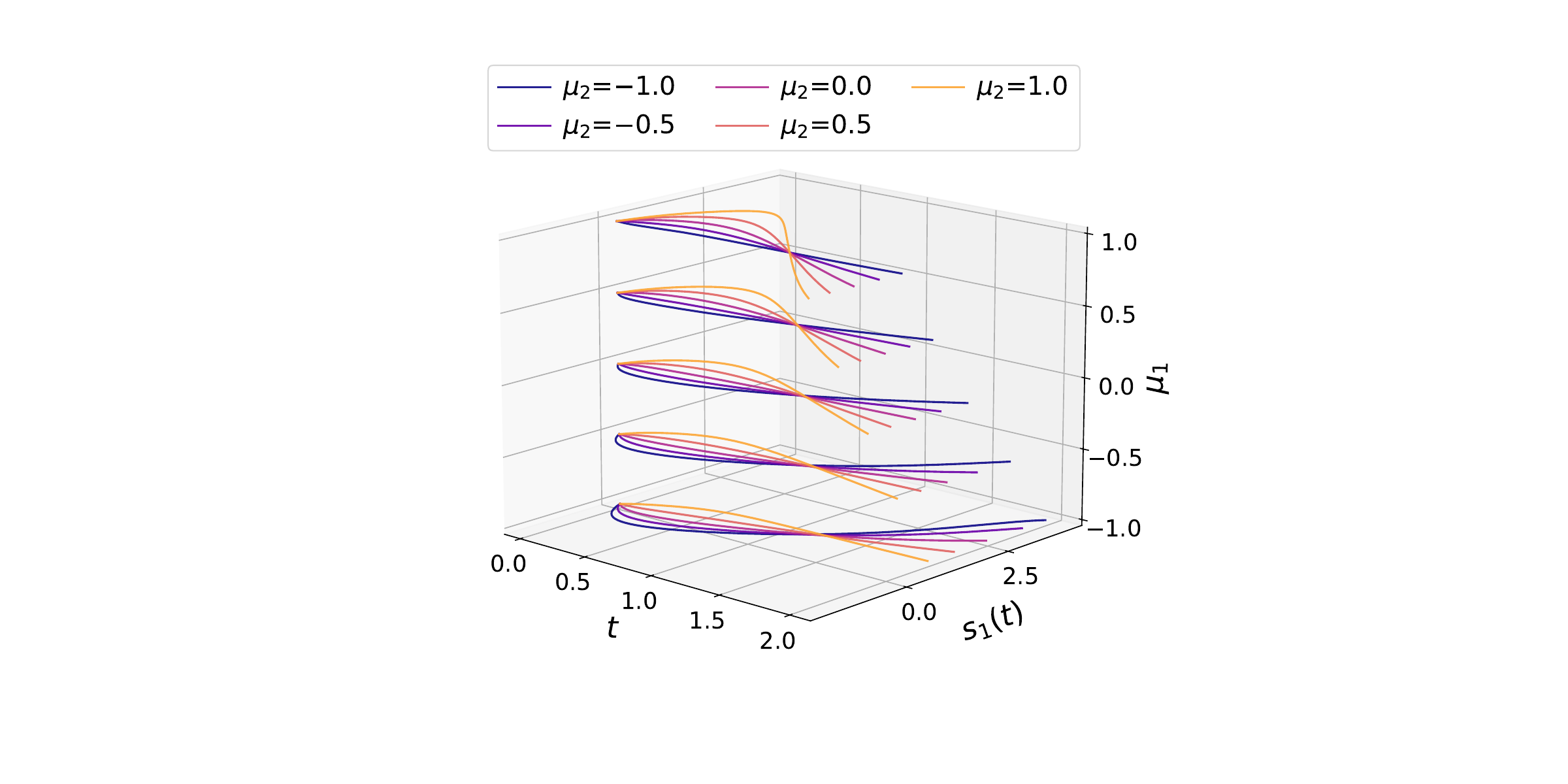}
        \caption{Evolution of $s_1$ for varying values of $\bm\mu$}
        \label{fig:latents_3d_1}
    \end{subfigure}
    \hfill
    \begin{subfigure}[b]{0.54\textwidth}
        \centering
         \includegraphics[width=\textwidth]{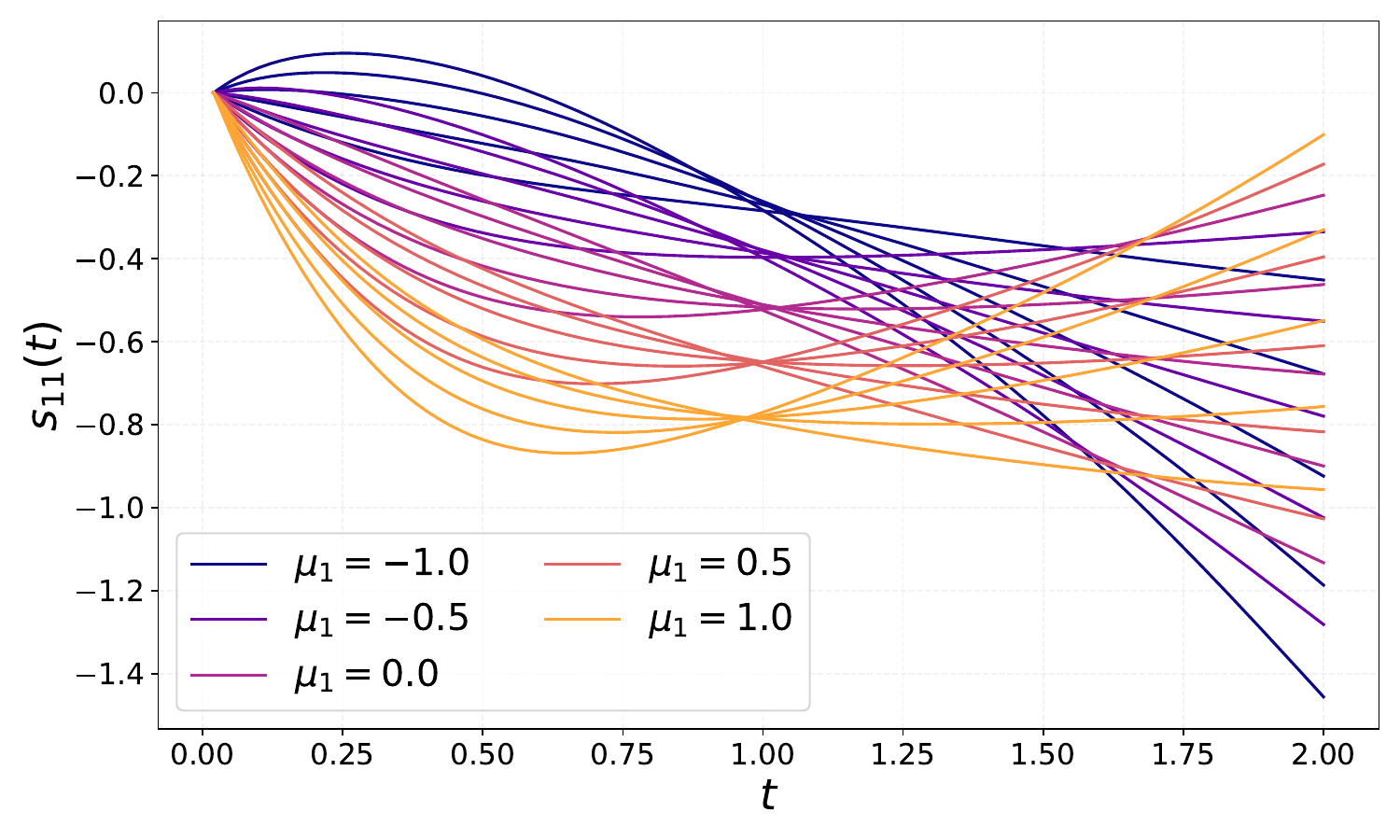}
        \caption{Evolution of $s_{11}$ for varying values of $\bm\mu$}
        \label{fig:sub_groups_2_SA_15}
    \end{subfigure}
    \caption{Behavior in time of two different components of $\bm{s}(t; \bm\mu)$ as $\bs{\mu}$ varies.} 
    \label{fig:latent_fixed_mu1mu2_SA}
\end{figure}

A first confirmation of such regularity comes from the plots in Figure \ref{fig:latent_fixed_mu1mu2_SA}, revealing clear and structured patterns for the latent trajectories.
Specifically, Figure \ref{fig:latents_3d_1} shows the evolution of the first component of $\bm s$ for $n=3$, exhibiting a consistent qualitative behavior across simulations corresponding to varying values of $\bm\mu$.

Similar considerations can be made in the case of $s_{11}$ in Figure \ref{fig:sub_groups_2_SA_15}, corresponding to a training setup with $n=15$.
The trajectories are clustered in the latent space according to the corresponding values of $\bs{\mu}$, with an interpretable evolution that can be easily detected and understood to obtain a more in-depth knowledge of the phenomenon, e.g., in the context of inverse problems and system identification tasks. 
In our simulations, increasing the latent dimension $n$ tends to enhance the interpretability of the trajectories.
For instance, in Figure \ref{fig:sub_groups_2_SA_15} it can be seen that, for all values of $\bm\mu_2$, the trajectories intersect at time $t=1$, which corresponds to the instant at which the convection term in Equation~\eqref{eq:SquareAdvStrong1} vanishes for each $\bm\mu\in\mP$.
This phenomenon occurs coherently for different simulations and, although the exact intersection time may slightly deviate from $t=1$, likely due to scaling effects in the latent dynamics, it highlights a strong and robust connection between the latent trajectories and the underlying system dynamics.

Having assessed the interpretability of the resulting latent trajectories, following the argument from Subsection \ref{subsec:interpolation}, we explore the possibility to perform zero-shot prediction, i.e., without integrating the system in-time, by means of latent states interpolation across parameters and time.
In particular, we employed two different interpolation strategies, \gls{gpr} \cite{rasmussen_gaussian_2005} and spline functions \cite{quarteroni_numerical_2007} of degree $1$.
To retain interpretability and given the low-dimensionality of the latent coordinates, a separate interpolant/regressor was trained for each latent dimension, considering the case $n=3$.
Remark that, among these two choices, the only one that allows us to extrapolate in time is using \gls{gpr}.
We then fitted the regressors/interpolators using the latent trajectories corresponding to training values.
We remark that we did not use the interpolants to perform extrapolation in time (see Remark \ref{rmk:interp_extrap} for more details), but we integrated the training trajectories up to time $t=2$ before performing interpolation.
The hyperparameters of the \gls{gpr} models are detailed in Appendix~\ref{sec:appendix}.

Now, following the procedure described in Subsection \ref{subsec:interpolation}, given a parameter value $\bs{\mu}$ and a time $t$, with $(t,\bs\mu)\in I\times\mP$, we can provide an estimate of the solution $\bm u_h(t; \bs{\mu})$ considering the interpolated latent state $\tilde{\bs s}(t; \bs{\mu})$ and decoding it using $\mathcal{NN}_\text{dec}$, thus obtaining $\bs{u}_\text{interp}(t; \bs{\mu})=\mathcal{NN}_\text{dyn}\left(t, \tilde{\bs s}(t; \bs{\mu}), \bs{\mu}\right)$,
in which $\text{interp}\in\{\text{GPR}, \,\text{spline}\}$.

\begin{figure}[htbp]
    \centering
    \begin{subfigure}[b]{0.59\textwidth}
        \centering
        \includegraphics[width=\textwidth]{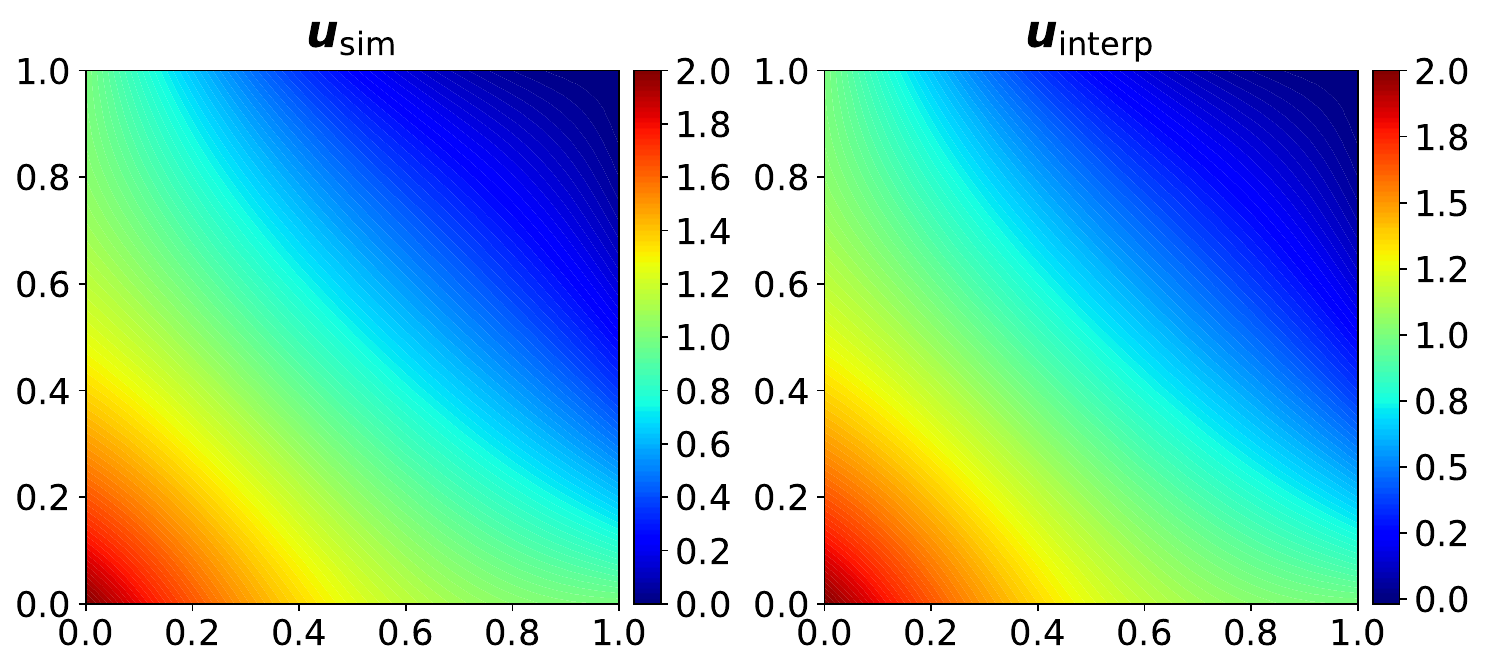}
        \caption{\gls{ldgcn} (left) and \gls{gpr}-interpolated (right) at $t=1$}
        \label{fig:interp_fields_SA}
    \end{subfigure}
    \hfill
    \begin{subfigure}[b]{0.4\textwidth}
        \centering
        \includegraphics[width=\textwidth]{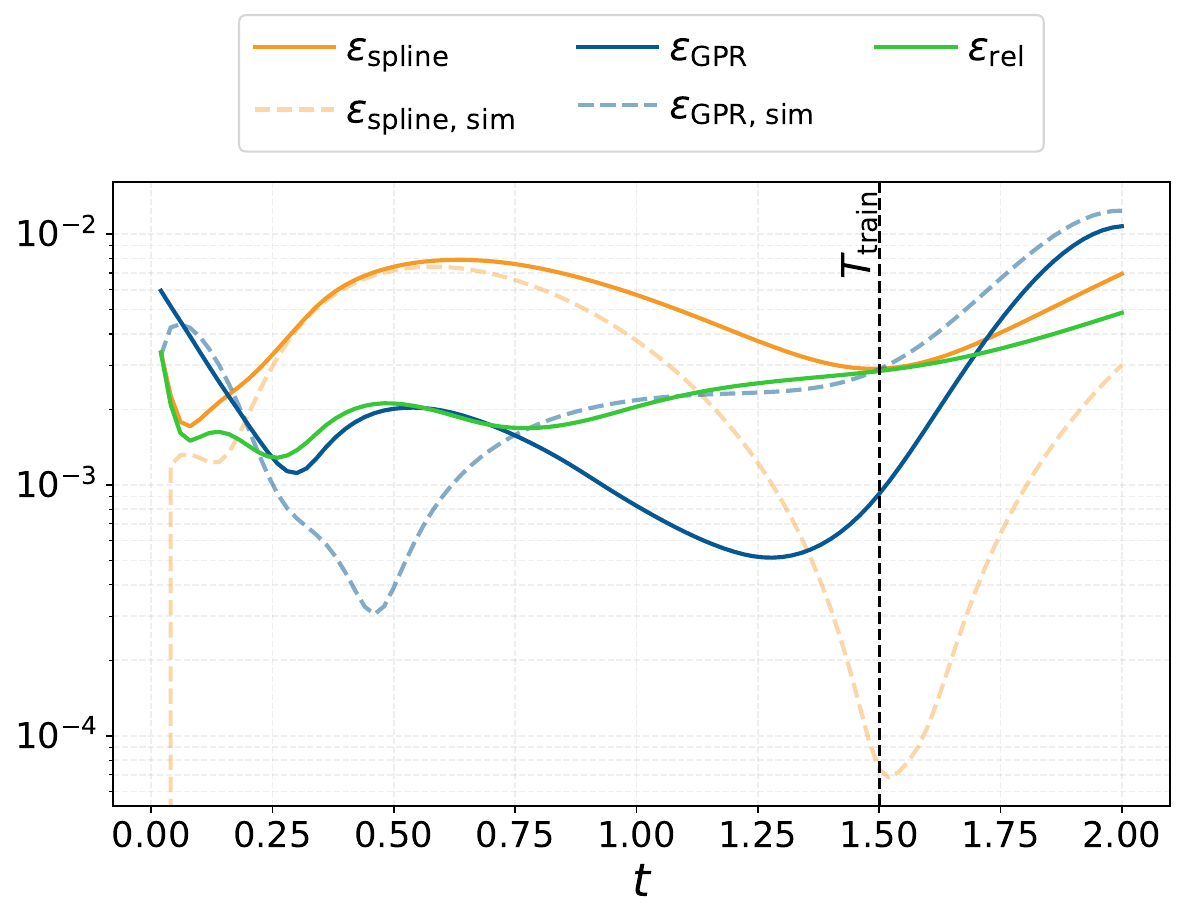}
        \caption{Comparison between relative errors}
        \label{fig:interp_errors_SA}
    \end{subfigure}
    \caption{Interpolated solution and errors for $\bs{\mu}_\text{test}=(0, 0)$.}
    \label{fig:interp_SA_error_field}
\end{figure}

We begin our investigation by showing in Figure \ref{fig:interp_fields_SA} a comparison between a decoded \gls{gpr}-interpolated field for the test value $\bs{\mu}_\text{test}=(0, 0)$ at time $t=1$ and its corresponding latent-integrated \gls{ldgcn} solution $\bs{u}_\text{sim}(t; \bs{\mu}_\text{test})$.
We emphasize that the parameter value $\bs{\mu}_\text{test} = (0,0)$ corresponds to the unique element of the dataset for which the advection term vanishes identically, providing the opportunity to test the quality of latent interpolation in a completely different dynamical regime.
Figure \ref{fig:interp_errors_SA} shows the behavior of \gls{ldgcn} via $\varepsilon_\text{rel}$, and of the relative errors of the interpolated strategies $\varepsilon_\text{interp}$ and $\varepsilon_\text{interp,sim}$ w.r.t.\ to $\bs u_h$ and $\bs u_\text{sim}$, respectively.
Indeed, from the theoretical consideration in Equation \eqref{eq:error_interp}, it follows that if one can bound with enough accuracy the error $\|\bs u_\text{sim}-\bs u_\text{interp}\|$, then $\bs u_h\approx \bs u_\text{interp}$.
From the figure we can observe that the errors using \gls{gpr} and spline interpolation are comparable, with \gls{gpr} returning smoother approximations and allowing to perform extrapolation.

\begin{figure}[htbp]
\begin{subfigure}[b]{0.49\textwidth}
     \centering
    \includegraphics[width=\textwidth]{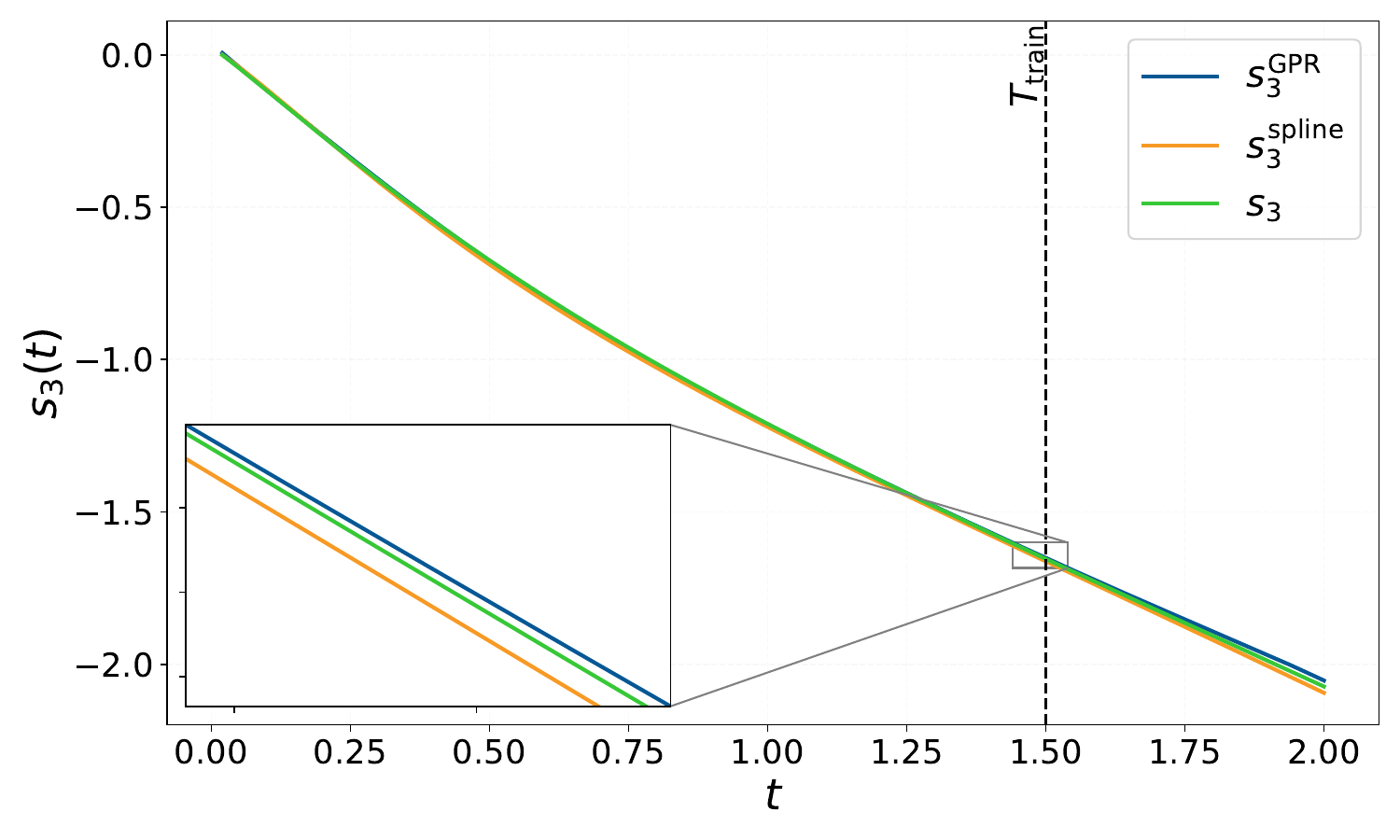}
    \caption{\textrm{Integrate-then-interpolate} strategy}
    \label{fig:interpolation_no_extrapolation}
\end{subfigure}
\begin{subfigure}[b]{0.49\textwidth}
    \centering
    
    \includegraphics[width=\textwidth]{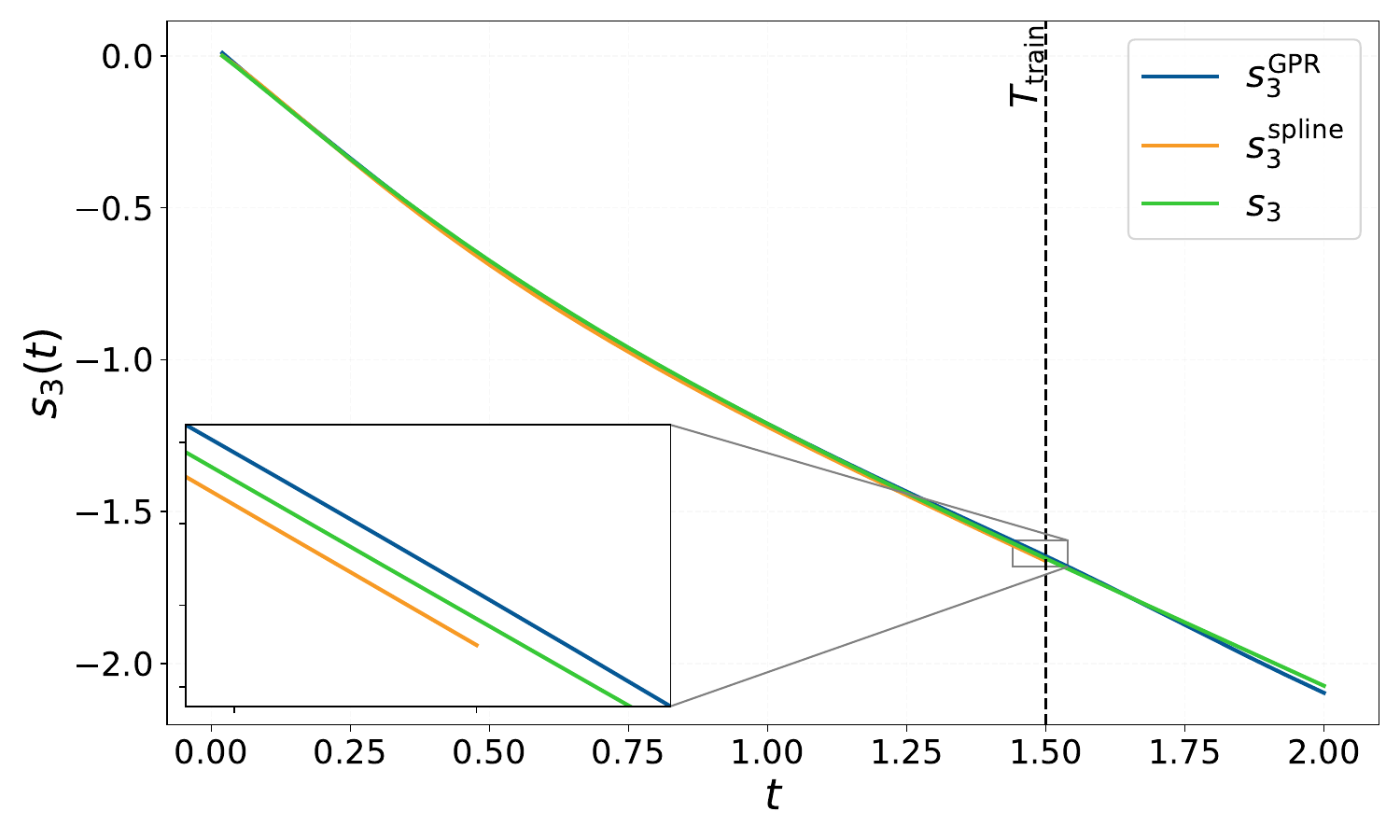}
     \caption{\textrm{Interpolate-then-extrapolate} strategy}
    \label{fig:interpolation_extrapolation}
\end{subfigure}
    \caption{Evolution of $s_3$ for both latent trajectories with $\bs{\mu}_\text{test}=(0, 0)$.}
    \label{fig:interp_latents_SA}
\end{figure}

Moreover, we plot in Figure \ref{fig:interp_latents_SA} the real and latent interpolated trajectories for $\bs\mu_\text{test}=(0, 0)$, obtained using the two strategies outlined in Remark \ref{rmk:interp_extrap}.
The figure shows that the two approaches yield similar predictions, with the main distinction being that spline functions cannot be used for extrapolation (Figure \ref{fig:interpolation_extrapolation}).
Although the trajectories show slight deviations during the extrapolation phase, the results remain satisfactory.

\begin{figure}[htbp]
    \centering
    \begin{subfigure}[b]{0.49\textwidth}
        \centering
        \includegraphics[width=\textwidth]{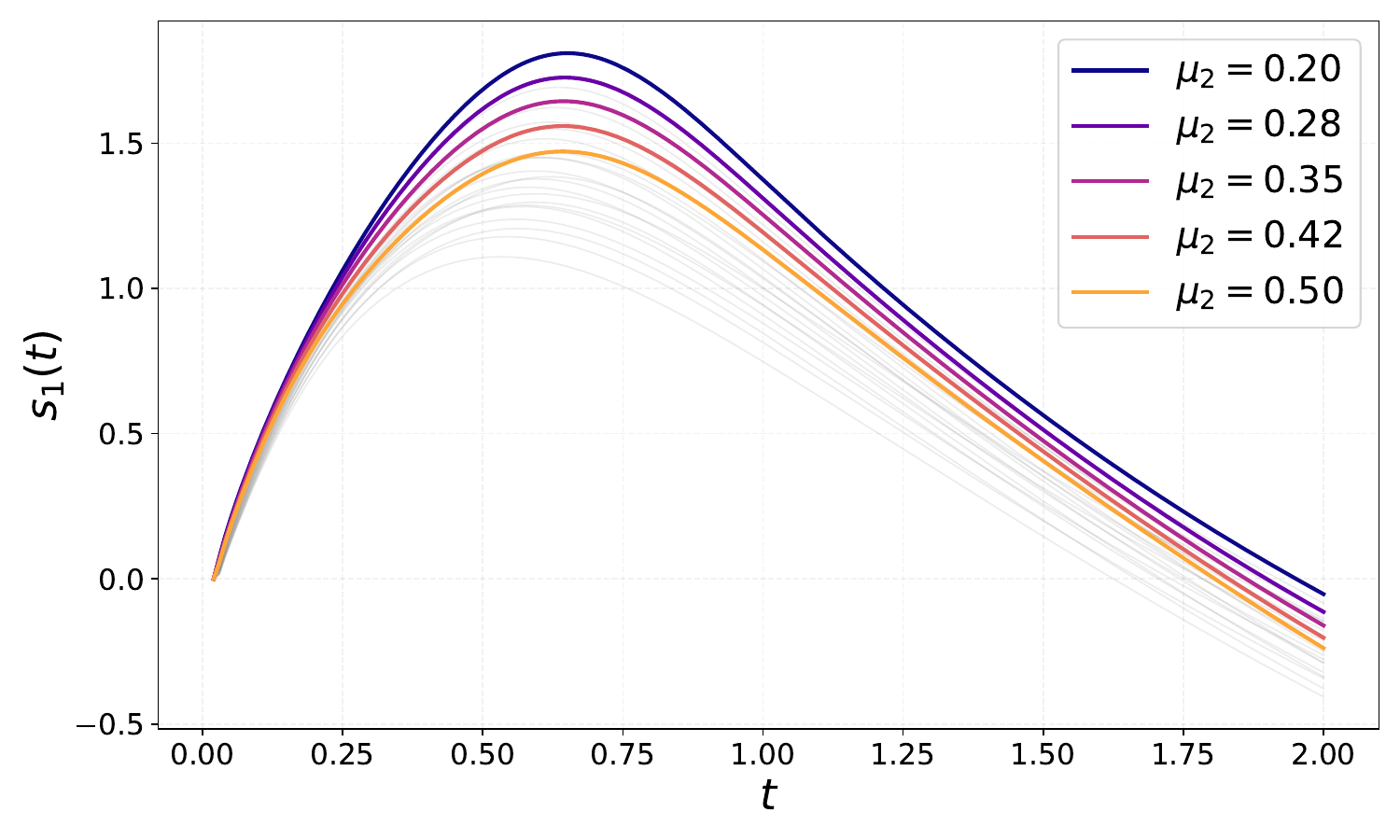}
        \caption{Evolution of $s_1$ for $\mu_1=0.2$}
        \label{fig:sub_groups_1_MH}
    \end{subfigure}
    \hfill
    \begin{subfigure}[b]{0.49\textwidth}
        \centering
        \includegraphics[width=\textwidth]{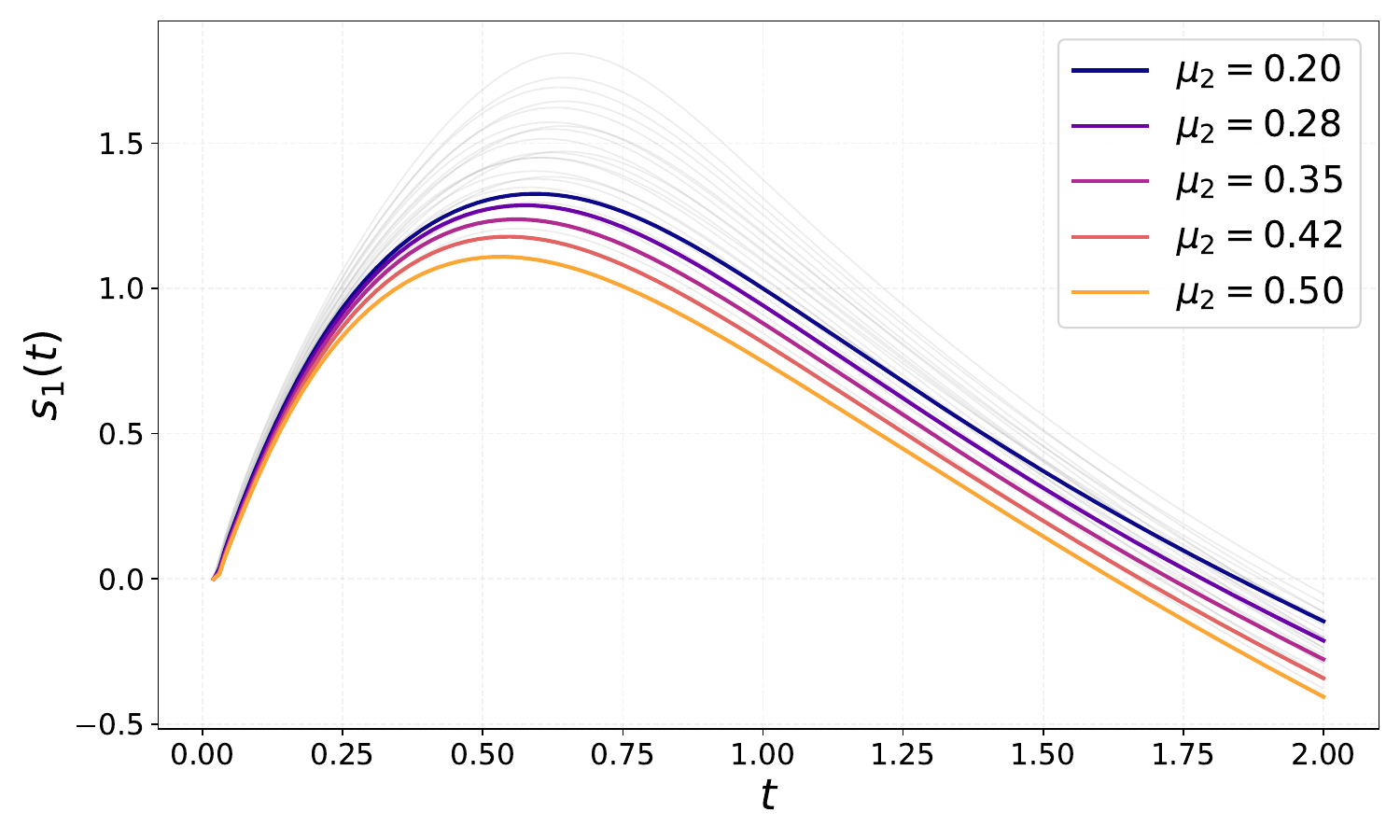}
        \caption{Evolution of $s_1$ for $\mu_1=0.5$}
        \label{fig:sub_groups_2_MH}
    \end{subfigure}
    \caption{Evolution of $s_1$ for two different fixed values of $\mu_1$ while varying $\mu_2$.}
    \label{fig:latent_grouped_MH}
\end{figure}

Finally, similar considerations can be made also in the more complex \gls{mh} benchmark with geometric parameters, where the trajectories exhibit regular patterns, as depicted in Figure \ref{fig:latent_grouped_MH} for $n=15$, where light gray lines correspond to other parameter instances, and the interpolation task can be easily performed to obtain efficient and global approximation of the dynamics.

The numerical results confirm that, as expected from the consideration in Proposition \ref{prop:interpolation}, all previous procedures are capable of accurately approximating solutions exploiting the robustness and interpretability of the learned latent dynamics exploiting interpolation strategies.
Nevertheless, we stress the fact that the performance might still suffer from degradation compared to the one of the \gls{ldgcn} architecture with time integration, suggesting that in more complicated scenarios the sole regularization induced in the latent space by the use of dynamical information might not be sufficient to understand the system dynamics when decoupled from the actual time integration.
Summarizing, when relying on the interpolation tool for zero-shot global approximation and flexibility, the causality of \gls{ldgcn} is lost and the feasibility of the method depends on the complexity of the problem and behavior of the latent space.
We anticipate that, for instance, Subsection \ref{subsec:coanda} provides an example in which the dynamics of the latent trajectories is less trivial due to the loss of well-posedness of the problem at the bifurcation point \cite{seydel_practical_2010,pichi_artificial_2023,pichi_graph_2024}.

\subsection{Lid-driven cavity flow}\label{subsec:lid_driven}
The third test case that we consider is a Navier--Stokes benchmark for flows inside a cavity, taking inspiration from the setup proposed in the original \gls{ldnet} paper \cite{regazzoni_learning_2024}.
This way, we could provide some insights on the performance of the architecture when compared to the original meshless architecture.

A key consideration in this setting is that \gls{ldgcn} is designed to exploit geometric information from complex domains -- a benefit that is diminished in the relatively simple setting of this benchmark.
This allows us to provide a fair comparison between the two strategies in which we are not exploiting any additional information, i.e., geometric inductive bias.
We remark that, given the discussion in Subsection \ref{subsec:ldgcn}, and the usual difficulty of comparing different machine learning architectures, the goal here is to highlight the capability of \gls{ldgcn} to retain the same good properties of \gls{ldnet} even when the task is not specifically suited for our framework.
Finally, for consistency with \cite{regazzoni_learning_2024}, in this section we employ the \gls{nrmse} metric to assess the quality of the reconstruction, defined as 
\begin{equation*}
    \text{NRMSE} = \frac{1}{u_\text{ref}}\sqrt{ \frac{1}{|\mathcal{T}_{h,\text{test}}| N_h d_{\bm u}} \sum_{(t,\bm\mu)\in \mathcal{T}_{h,\text{test}}} \sum_{i=1}^{N_h} \sum_{k=1}^{d_{\bm u}}\big( u_{h}^{i,k}(t;\bm\mu) - u_{\text{sim}}^{i,k}(t;\bm\mu) \big)^2},
\end{equation*}
with $u_\text{ref}\in\R$ a reference value.
We underline that, by averaging the space, time, and parameter variables, \gls{nrmse} makes it more difficult to spot large but localized errors.

\subsubsection{Problem definition and data collection}
Let us consider the domain $\Omega = (0,1)^2$, we are interested in solving the incompressible Navier--Stokes equations:
\begin{equation}\label{eq:NS}
\begin{cases}
\rho \dfrac{\partial \bs{u}}{\partial t}(\bm x,t) + (\bs{u}(\bm x,t) \cdot \nabla)\bs{u}(\bm x,t) = -\frac{1}{\rho}\nabla p(\bm x,t) + \mu \nabla^2 \bs{u}(\bm x,t), & \forall(\bm x, t)\in \Omega \times (0, T], \\
\nabla \cdot \bs{u}(\bm x, t) = 0, & \forall(\bm x, t)\in \Omega \times (0, T],
\end{cases}
\end{equation}
where $\bs{u} = [u_x, u_y]^T$ is the velocity field, $\rho$ is the fluid density, $\mu$ the kinematic viscosity, and the final time is $T=2$.
To ensure well-posedness of Equations \eqref{eq:NS}, we assume that suitable boundary and initial conditions are imposed as:
\begin{equation}\label{eq:BC_s_NS_LD}
\begin{cases}
\bs{u}(\bm x,t) = v(t)\, \bs{e}_x, & \forall(\bm x, t)\in\Gamma_{\text{top}} \times (0, T], \\
\bs{u}(\bm x,t) = \bs{0}, & \forall(\bm x, t)\in\partial\Omega \setminus \Gamma_{\text{top}} \times (0, T], \\
\bs{u}(\bs x, 0) = \bs{0}, & \forall \bs x \in \Omega,
\end{cases}
\end{equation}
in which $\Gamma_{\text{top}}$ is the top edge of the domain, where a time-dependent boundary condition defined as $v(t) = \sum_{k=1}^{N_F} \frac{\mu_k}{k^2} \sin\left( \frac{2\pi k t}{T} \right)$ is imposed, with $N_F$ the number of Fourier modes, and $\mu_k$ the coefficients determining the shape of $v(t)$ across simulations.
To keep the maximum Reynolds number around $850$, the values for $\mu_k$ are sampled from a Gaussian distribution with mean $\mu = 0$ and standard deviation $\sigma = 3$.
We remark that this particular parameterization, encoded in the boundary conditions, creates a challenging benchmark which shares similarities with the operator learning task.

To generate the dataset and investigate the use of \gls{ldgcn} to predict the two components of the velocity solution, an unstructured mesh was employed, consisting of a total of $\num{10024}$ nodes using the Taylor--Hood pair $\mathcal{P}^2/\mathcal{P}^1$ to satisfy the \emph{inf-sup} condition \cite{quarteroni_numerical_2017}.
The kinematic viscosity was set to $\mu = 10^{-2}$, while the density was fixed at $\rho = 1$.
The time discretization was performed using a timestep $\Delta t = 10^{-1}$ with the implicit Euler method \cite{quarteroni_numerical_2007}.
We considered $80$ different realizations of $v(t)$, and a total of $19$ velocity snapshots for each trajectory. Indeed, we removed the first snapshots (being equal to zero) and the last one (due to periodicity).

\subsubsection{Performance of the LD-GCN architecture}
We trained a \gls{ldgcn} model by randomly selecting $80\%$ of the trajectories, from which we also excluded the last $50\%$ of the evolution in time, resulting in the training interval $I_{h,\text{train}}=[0.1, 0.9]$.
In this setup, $576$ out of the total $\num{1520}$ snapshots were seen by the network during training, allowing us to test its extrapolation properties.

By fixing the latent dimension to $n=3$, we obtained a \gls{nrmse} on the test set equal to $6.79\cdot 10^{-3}$, confirming the great performance in reconstructing the dynamics even in such scarce-data regime with non-trivial parameterization and long-time extrapolation.

Moreover, the results are fairly consistent with the those presented in \glspl{ldnet}, showing the same order of magnitude (reported to be $1.39\cdot10^{-3}$ using $n=10$).
Table \ref{tab:err_LD} compares the errors obtained using our methodology for varying values of $n$.

It can be observed that, in this particular case, the amount of weights for the  \gls{ldgcn} architecture is elevated.
As discussed in Subsection \ref{subsec:ldgcn}, this is due to the presence of the fully-connected layers that are used to learn the node embeddings in the graph convolutional decoder.
We remark that, since the output of the last fully-connected layer has size $N_h\times d_{\bm u}$, the number of trainable weights could potentially limit the applicability of \gls{ldgcn} to the case of very fine meshes.
Nevertheless, the architecture was still able to obtain accurate predictions even with $N_h \sim 10^4$ and $d_{\bm u} =2$.
Potential strategies to alleviate the problem are described in Section \ref{sec:conclusions_and_perspectives}.

\begin{table}[htbp]
    \caption{\gls{nrmse} and number of parameters of \gls{ldgcn} for varying values of $n$. The last two columns refer to the number of weights of each subnetwork.}
    \centering
    \begin{tabular}{|c|c|c|c|}
        \hline
        \rowcolor{white}
      $n$ & \gls{nrmse} & $\NN{dyn}$ & $\NN{dec}$\\ 
        \hline
      \cellcolor{gray!20}$3$ &\cellcolor{gray!20} $6.79\cdot10^{-3}$ &\cellcolor{gray!20}$\num{21543}$ & \cellcolor{gray!20}$\num{4030884}$ \\
          \cellcolor{gray!5}$10$ &\cellcolor{gray!5} $8.33\cdot10^{-3}$ &\cellcolor{gray!5}$\num{22250}$ &\cellcolor{gray!5}$\num{4032284}$\\
        \cellcolor{gray!20}$15$ &\cellcolor{gray!20} $6.84\cdot10^{-3}$ &\cellcolor{gray!20}$\num{22755}$ &  \cellcolor{gray!20}$\num{4033284}$ 
        \\
        \hline
    \end{tabular}
    \label{tab:err_LD}
\end{table}

\begin{figure}[htbp]
    \centering
    \includegraphics[width=.48\textwidth]{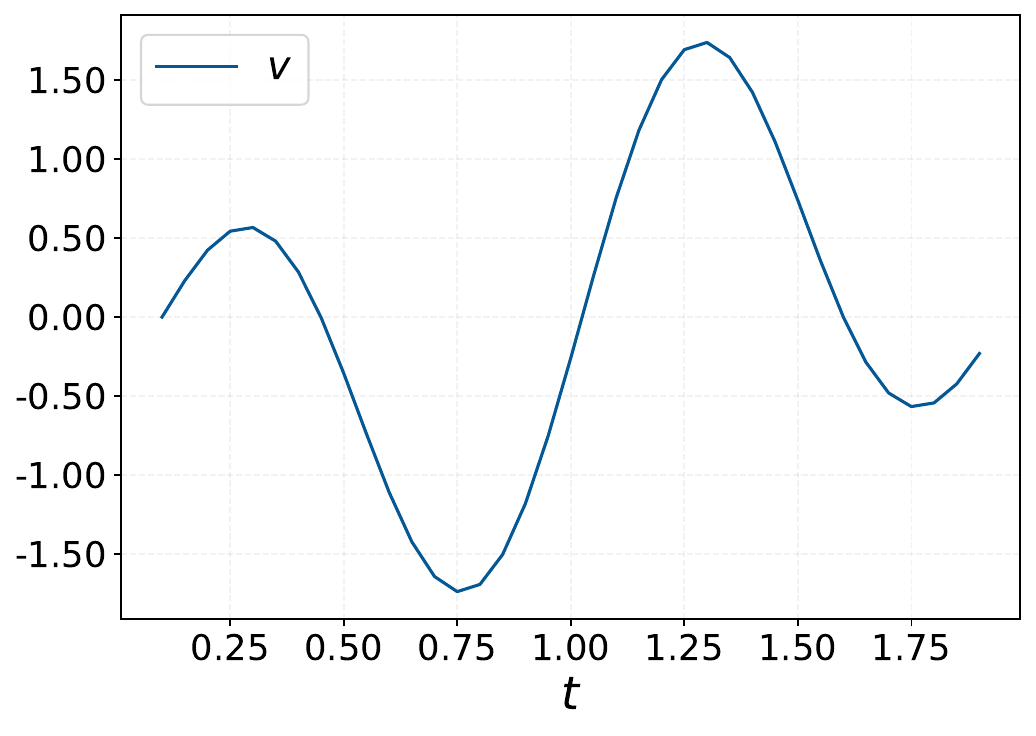}
    \includegraphics[width=.48\textwidth]{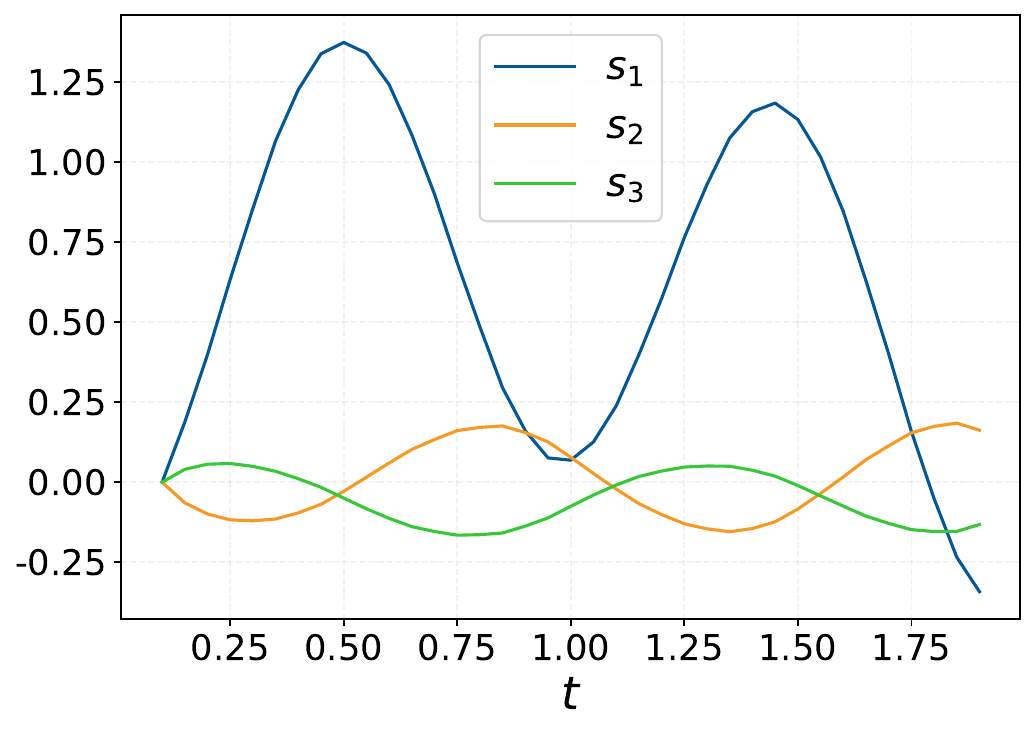}
    \caption{Realization of the signal $v(t)$ and corresponding trajectory $\bs s(t; \bs{\mu})$.}
    \label{fig:latents_ld}
\end{figure}

Figure \ref{fig:latents_ld} shows a realization of the signal $v(t)$ and the corresponding latent trajectory, highlighting that the latter mirrors the oscillatory behavior of the former.

\begin{figure}[htbp]
    \centering
    \begin{subfigure}[b]{\textwidth}
        \centering
        \includegraphics[width=\textwidth]{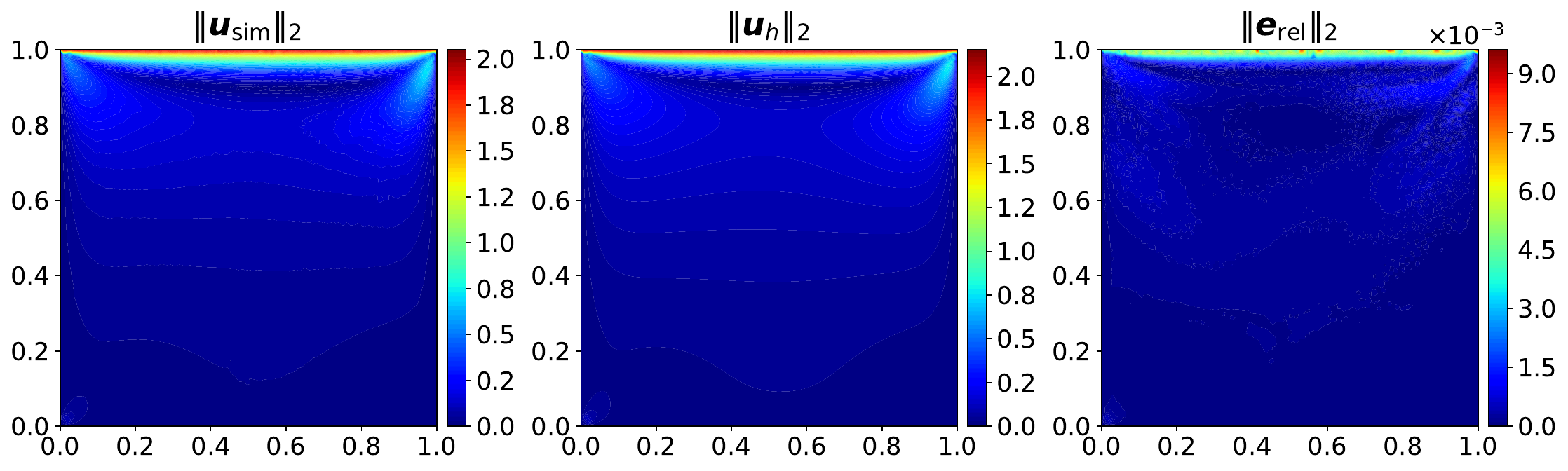}
        \caption{$t=0.1$}
    \end{subfigure}
    \hfill
    \begin{subfigure}[b]{\textwidth}
        \centering
        \includegraphics[width=\textwidth]{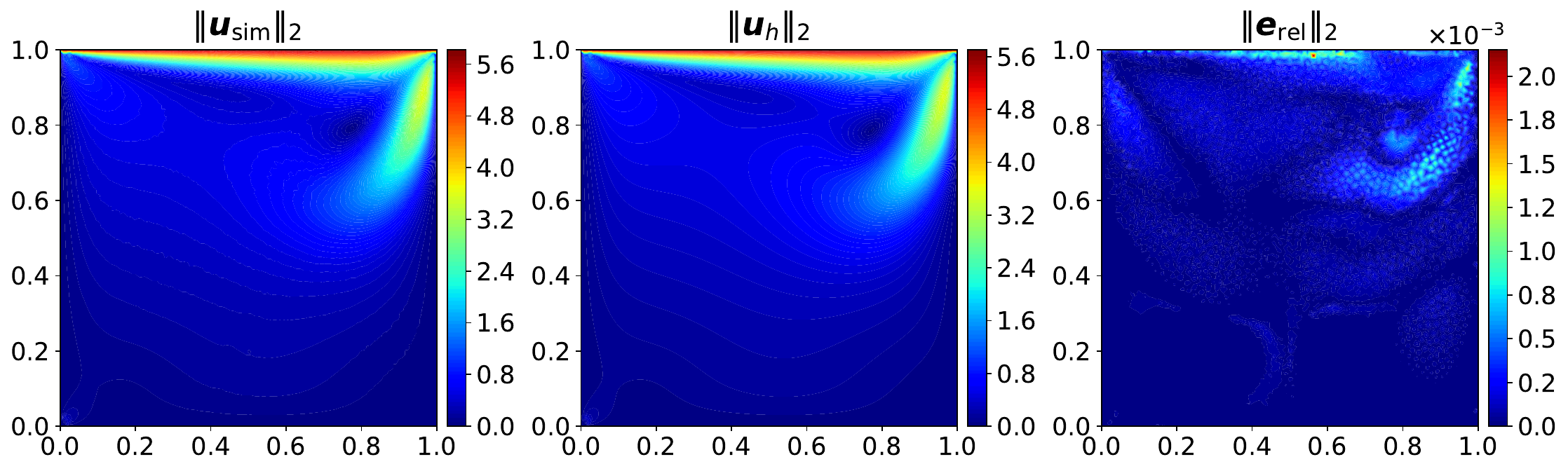}
        \caption{$t=0.6$}
    \end{subfigure}
    \caption{Plot of the norm of the simulated solution (left), of the full-order one (middle) and of the relative $L^2$ error (right) for a given signal.}
    \label{fig:ld_trajectories}
\end{figure}

To conclude, Figure \ref{fig:ld_trajectories} shows a few simulated snapshots and the comparison with the full-order ones.
It can be seen that the transition from the inlet boundary layer to the formation vortex pattern is correctly recovered with high accuracy, so that even the level curves closely match those of the full-order model.

\subsection{Coand\u{a} effect}\label{subsec:coanda}
Finally, the last benchmark is concerned with the analysis of a bifurcating phenomenon in fluid dynamics.
Bifurcations are characterized by the coexistence of multiple solutions for the same parametric values, so that small perturbations around critical values lead to sudden qualitative changes in system dynamics \cite{dijkstra_bifurcation_2023, seydel_practical_2010}.

Here, we are not interested in reconstructing all branches of solutions coexisting for each parameter value.
Instead, our goal is to assess the performance of \gls{ldgcn} for a highly complex and potentially ill-posed task, and to investigate whether the learned trajectories retain information about the underlying bifurcation, i.e., if the bifurcation can be detected or discovered already at the latent level.

\subsubsection{Problem definition and data collection}
The Coand\u{a} effect is related to wall-hugging and symmetry-breaking behaviors for channel flows, and represents an important benchmark to study the lack of uniqueness for fluid simulations, especially concerning the stability properties when increasing the Reynolds number.
In particular, we consider the setting presented in \cite{tomada_sparse_2025,pichi_artificial_2023, pichi_driving_2022}, to which we refer for a more detailed description and investigation of the bifurcating phenomenon.

We denote by $\Omega$ the interior of the domain $\bigl([0, 10]\times [2.5, 5]\bigr)\cup \bigl([10, 50]\times[0, 7.5]\bigr)\subset\R^2$ representing a sudden-expansion channel.
Our objective is to predict the velocity field $\bm u$ of the solution to the Navier--Stokes equations \eqref{eq:NS}, subject to the following initial and boundary conditions:
\begin{equation}\label{eq:bc_coanda}
    \begin{dcases}
    \bs{u}(\bs{x}, 0) = \bs{0}, &\forall \bm x \in\Omega, \\
    \bs{u}(\bs{x}, t) = [20(y - 2.5)(5 - y),\, 0], &\forall(\bm x,t)\in\Gamma_{\text{in}}\times(0,T), \\
    \bs{u}(\bs{x}, t) = \bs{0}, &\forall (\bm x ,t)\in\Gamma_0\times(0,T), \\
   \bigl (-\mu\nabla\bs{u}(\bm x,t) + p(\bm x,t)\mathbf{I}\bigr)\bs{n}(\bm x) = \bs{0}, &\forall(\bm x, t)\in\Gamma_{\text{out}}\times(0,T),
    \end{dcases}
\end{equation}
where $\mu$ is the kinematic viscosity, and the boundaries are defined as $\Gamma_{\text{in}}=\{0\}\times [2.5,\,5]$, $\Gamma_{\text{out}} = \{50\}\times[0,\, 7.5]$, and $\Gamma_0 = \partial\Omega\smallsetminus(\Gamma_{\text{in}}\cup\Gamma_{\text{out}})$. Since we are interested in investigating the system dynamics near the critical parameter, i.e., the bifurcation point, known for this setting to be around $\mu^\ast \approx 0.96$, we considered the kinematic viscosity as parameter varying in the range $\mu \in \mP=[0.8, 1.2]$.

To generate the training and testing data, we performed numerical simulations for $20$ equispaced values of $\mu$.
Time integration was performed using the trapezoidal rule with $\Delta t=10^{-2}$ until the final time $T=120$, while the domain $\Omega$ was discretized using a mesh with $2752$ nodes over which we solved the Navier--Stokes equations with the \gls{fe} inf-sup stable Taylor--Hood $\mathcal{P}_2/\mathcal{P}_1$ pair.
Since the early transient does not provide information relevant to the bifurcation behavior \cite{tomada_sparse_2025}, describing the evolution of the inlet profile in the ``empty'' channel, we only retained data for $t\geq t_0=8$, and sub-sampled with a step of $0.5$ to consider again a scarce-data scenario.

\subsubsection{Model performance}
To study the in-time and in-parameter extrapolation properties also in this more complex setting, we chose a training/test ratio of 90\%/10\% for viscosity values and (last) time instances.
We trained a \gls{ldgcn} model with the same strategy outlined in Algorithm \ref{alg:ld_gcn}, with the goal to both predict the velocity field $\bm u$, and to investigate the features of the encoded latent variables representing the underlying bifurcation structure.

\begin{figure}[htbp]
    \centering
    \begin{subfigure}[b]{0.55\textwidth}
    \centering
    \includegraphics[width=\linewidth]{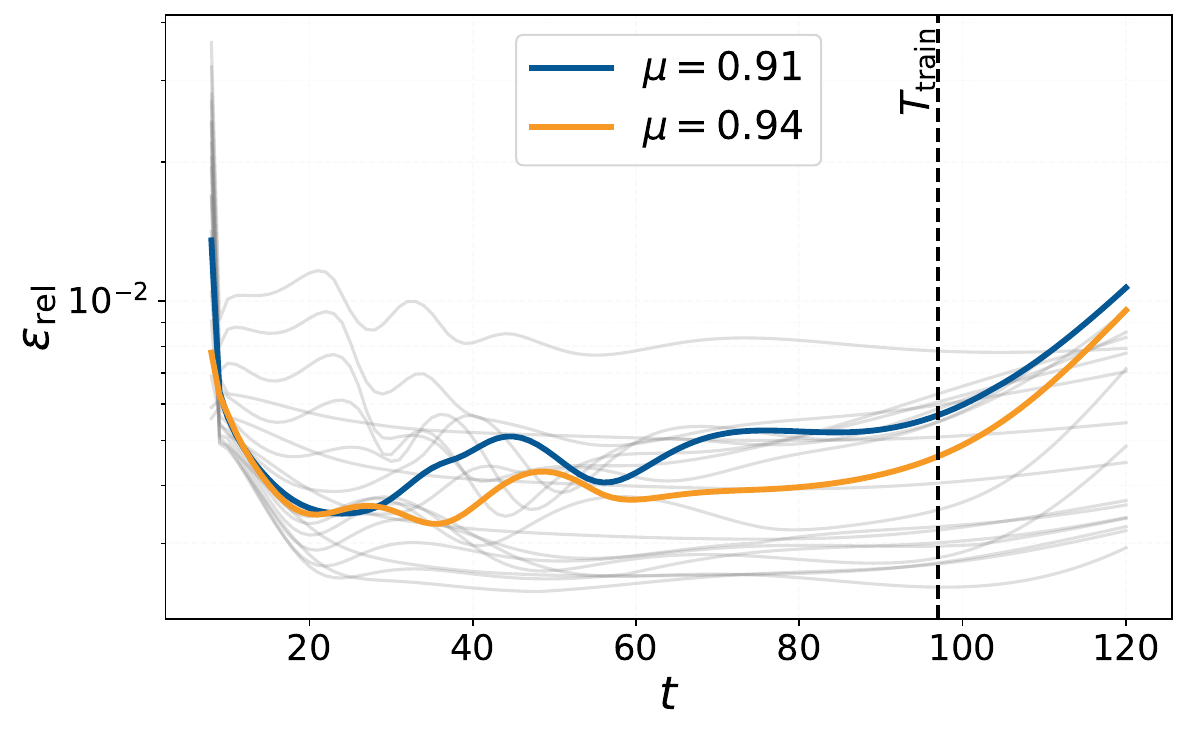}
    \caption{Relative errors}
    \label{fig:err_coanda}
    \end{subfigure}
    \hfill
     \begin{subfigure}[b]{0.44\textwidth}
        \centering
    \includegraphics[width=\linewidth]{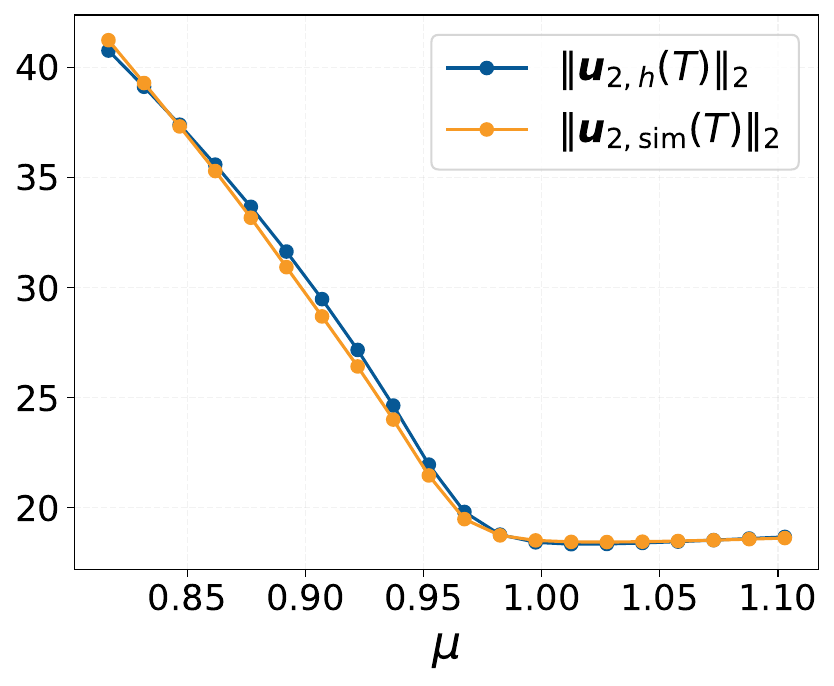}
    \caption{Bifurcation diagram}
    \label{fig:coanda_norm}
    \end{subfigure}
    \caption{Relative errors of \gls{ldgcn} for $\mu\in\mP_h$ and bifurcation diagram for \gls{fe} and \gls{ldgcn} approximations.}
    \label{fig:coanda_err_and_bif}
\end{figure}

We start by discussing the plot in Figure \ref{fig:err_coanda} showing the relative errors $\varepsilon_\text{rel}$ for the \gls{ldgcn} architecture obtained for all parameter values, highlighting the ones corresponding to the test (post-bifurcating) configurations, and using gray lines for values of $\mu\in\mP_{h,\text{train}}$.
Also in this setting, the model shows great accuracy, even in the extrapolation regime, by obtaining a maximum relative error of $\varepsilon_\text{max}=3.62\cdot10^{-2}$ and a mean relative error of $\varepsilon_\text{mean}=4.57\cdot10^{-3}$.

\begin{figure}[htbp]
    \centering
    \begin{subfigure}[b]{\textwidth}
        \centering
        \includegraphics[width=\textwidth]{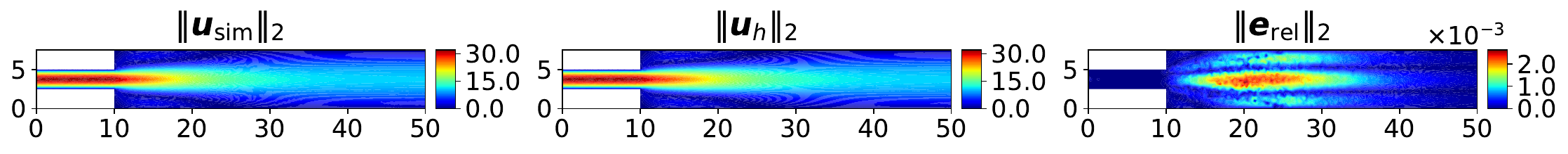}
        \caption{$\mu=1.1,\,t=8$}
        \label{fig:coanda_max_err}
    \end{subfigure}
    \hfill
    \begin{subfigure}[b]{\textwidth}
        \centering
        \includegraphics[width=\textwidth]{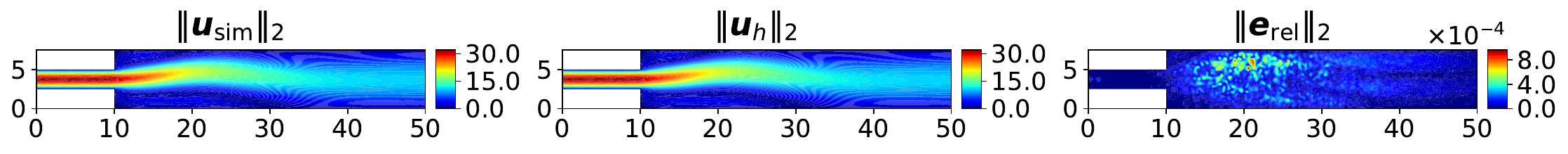}
        \caption{$\mu=0.82,\,t=120$}
        \label{fig:coanda_wall_hugging}
    \end{subfigure}
    \caption{Magnitude of the velocity field for pre- and post-bifurcating behaviors, top and bottom, respectively.}
    \label{fig:coanda_fields}
\end{figure}

Finally, in Figure \ref{fig:coanda_fields} we show two representative snapshots of the velocity field, highlighting the complexity of the chosen benchmark with the two qualitatively different behaviors admitted by the system, i.e., the symmetric pre-bifurcation configuration for $\mu = 1.1$, and the asymmetric wall-hugging one past the bifurcation point for $\mu = 0.8$. 
The former state, depicted in Figure \ref{fig:coanda_max_err}, corresponds to the maximum relative error $\varepsilon_\text{max}$ in the whole dataset, and being attained at the first time instance considered, it confirms the choice of excluding the initial ``mostly diffusive'' regime that does not bring any substantial information on the bifurcation detection. We remark that the stable bifurcating state in Figure \ref{fig:coanda_wall_hugging} is coherent with the one obtained via the \gls{fe} method, but we postpone to successive investigations the question concerning the possibility to recover multiple coexisting solutions and how their latent trajectories are characterized \cite{PichiDeflationbasedCertifiedGreedy2025a}.

\subsubsection{Bifurcation analysis}
We conclude the analysis by investigating whether the interpretability and robustness of the \gls{ldgcn} architecture allow to detect the bifurcating behavior of the system from the latent dynamics representation.

We start by depicting the so-called (pitchfork) bifurcation diagram in Figure \ref{fig:coanda_norm}.
Here, we exploit the norm of the vertical velocity $\bm{u}_{2,\ast}(T)$ of both the high-fidelity and \gls{ldgcn} approximations at final time $T$ as the quantity of interest to observe the predicted qualitative change when decreasing the viscosity value~\cite{pichi_artificial_2023,tomada_sparse_2025}. 
The model correctly reproduces the diagram and, by rapidly deviating from the symmetric configuration, accurately identifies the bifurcation point around $\mu^\ast\approx 0.96$.

\begin{figure}[htbp]
    \centering
    \begin{subfigure}[b]{0.48\textwidth}
        \centering
        \includegraphics[width=\textwidth]{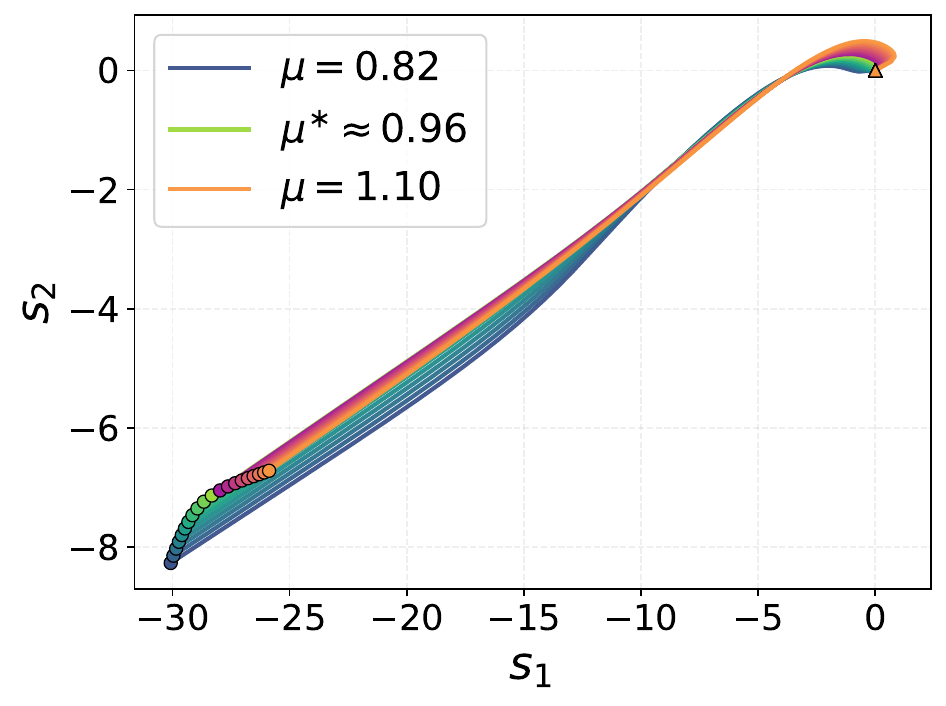}
         \caption{Phase portrait in the $(s_1,s_2)$ plane}
        \label{fig:coanda_phase_plane}
    \end{subfigure}
    \hfill
    \begin{subfigure}[b]{0.48\textwidth}
        \centering
        \includegraphics[width=\textwidth]{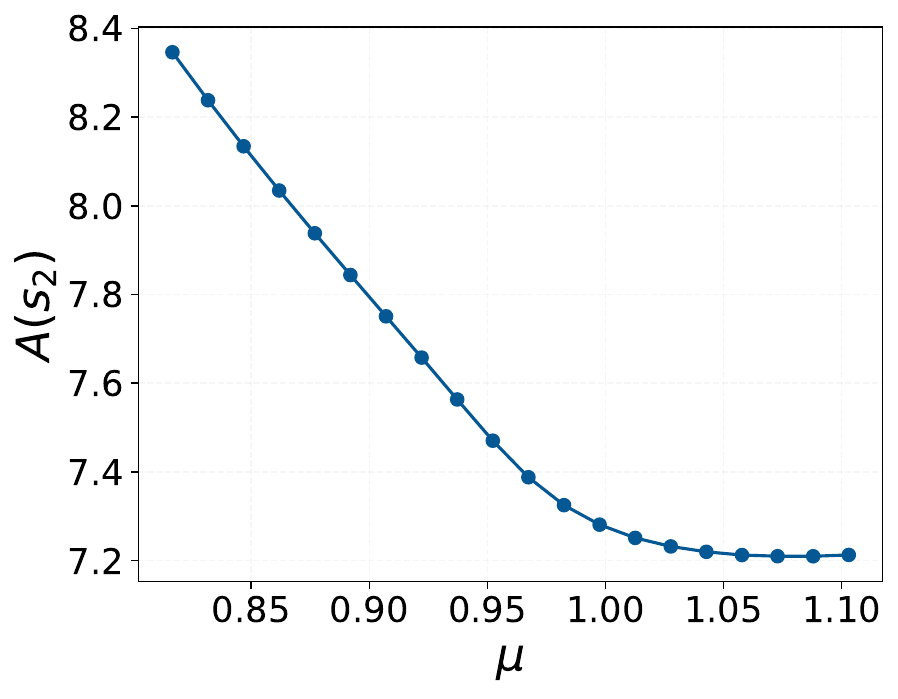}
        \caption{Amplitude $A(s_2)$}
        \label{fig:coanda_amplitude}
    \end{subfigure}
    \caption{Phase portrait and amplitude of latent trajectories. In the left plot, the triangle corresponds to the initial condition $\bm s(t_0;\,\cdot\,)=\bm 0$, while the dots denote the final values $\bm s(T;\,\cdot\,)$.}
    \label{fig:coanda_latent_bif_diagrams}
\end{figure}

To investigate the latent-space dynamics, Figure \ref{fig:coanda_phase_plane} shows the phase portrait of the system in the $(s_1, s_2)$ plane for $\mu\in\mP_h$.
While the latent trajectories do not exhibit sharp gradients or high-order discontinuities, thanks to the regularizing effect of the latent integration and dynamics, they nonetheless encode meaningful information about the transition from symmetric to asymmetric states.
In particular, trajectories corresponding to values in the bifurcation regime with $\mu \leq \mu^\ast$ tend to diverge, after an initial transient, from those associated with symmetric solutions.
Interestingly, this behavior mirrors that of the full-order model: even for asymmetric solutions, the wall-hugging behavior emerges only after an initially symmetric phase \cite{tomada_sparse_2025}.

Motivated by this behavior, we show in Figure \ref{fig:coanda_amplitude} the amplitude $A(s_2) = \max_{i\in\{0,\dots,N_t\}}s_2(t_{i})\;\;-\,\min_{i\in\{0,\dots,N_t\}}s_2(t_{i})$ computed for each $\mu\in\mP_h$.
Analyzing this quantity of interest yields a diagram that exhibits the same qualitative behavior observed in Figure \ref{fig:coanda_norm}, thereby enabling the construction of a bifurcation diagram based solely on the latent trajectories.

These results demonstrate that the latent dynamics learned by \gls{ldgcn} not only enable accurate prediction of the flow behavior, but also inherently encode the qualitative features of the bifurcation, without requiring access to full-order fields.
\section{Conclusions and Perspectives}\label{sec:conclusions_and_perspectives}
In this work, we introduced \gls{ldgcn}, an architecture suitable for non-intrusive \gls{mor} of time-dependent parameterized \glspl{pde} on unstructured meshes.
We showed that latent dynamics modeling can be successfully coupled with \glspl{gnn} for field reconstruction, allowing to efficiently study systems that depend on a global latent state, while leveraging the power of \glspl{gnn} for parameterized and varying geometries.
In addition, we investigated the latent space identified by the network, successfully interpolating the latent trajectories for better performance and interpretability.
Finally, we created a link with current state-of-the-art methodologies and gave a mathematical justification of the proposed strategy by proving a \gls{uat} for encoder-free architectures.

We evaluated our architecture in four challenging scenarios for \gls{mor} benchmarks and compared the results with state-of-the-art methodologies such as \gls{ldnet} and \gls{gca}, of which our approach can be seen as an extension.

Direct comparisons with \gls{gca} showed that \gls{ldgcn} achieves significant improvements in effectiveness in the two tests involving advection-dominated problems with physical and geometric parameters, especially when dealing with time extrapolation even for unseen parameter values.
Moreover, our approach reduced the number of trainable parameters by approximately $50\%$, avoiding overfitting, learning simple and interpretable latent states that are dynamically meaningful, and allowing for an even more lightweight online phase by supporting zero-shot prediction.

Similarly, in the lid-driven cavity flow scenario, \gls{ldgcn} achieved performance comparable to \gls{ldnet}, while for the Coand\u{a} effect it allowed to detect the  bifurcation diagram directly by investigating the latent dynamics.

Future work will focus on further extending \glspl{ldgcn} to overcome the main current limitations, namely the relatively large number of trainable parameters and the reliance on fixed mesh connectivity and cardinality.
Addressing these issues, for example by integrating a \gls{node} with \glspl{gfn} \cite{morrison_gfn_2024}, would allow to operate also in the multi-fidelity setting, where data could potentially come from different mesh discretizations. 
Another important direction is the extension of the methodology to an operator learning framework.
Such a development would enable modeling of systems with varying initial conditions and forcing fields.
Finally, developing error estimates and approximation theorems similar to those in \cite{farenga_latent_2025} also in case of time-varying signals would be important to provide even more general theoretical contribution.

\section{Acknowledgments}
The authors acknowledge the support provided by INdAM-GNCS and the European Union - NextGenerationEU, in the framework of the iNEST - Interconnected Nord-Est Innovation Ecosystem (iNEST ECS00000043 - CUP G93C22000610007) consortium and its CC5 Young Researchers initiative. 
The authors also acknowledge the early stage of development and discussion provided by Francesco Sala, Mariella Kast, and Gaspare Li Causi.

\renewcommand{\bibfont}{\small}
\printbibliography[]
\begin{appendix}
\section{Architectures}\label{sec:appendix}
Here, we discuss some details of the architectures exploited for the benchmarks considered in the manuscript.
The decoder $\NN{dec}$ coincides with that of the \gls{gca} architecture.
In particular, it consists of two convolutional layers based on MoNet \cite{monti_geometric_2017} with $n_{\text{hc}}=2$ hidden channels each, preceded by a two-layer neural network with shape $[n,\,200,\,N_h\times n_{\text{hc}}]$.
For all \gls{gca} models, the encoder has a symmetric architecture.
We refer the interested reader to \cite{pichi_graph_2024} for further details on the \gls{gca} architecture.

Except for the Coand\u{a} test case in Subsection \ref{subsec:coanda}, in all benchmarks $\NN{dec}$ also depends explicitly on $(t,\bs\mu(t))$ in view of Remark \ref{rmk:signature}.
There are two main reasons for this choice: firstly, it often results in better network performance; secondly, in the numerical experiments of Section~\ref{sec:results} the initial condition changes slightly for each instance of $\bm\mu$.
Since the latent initial condition is fixed, the only way to account for different initial conditions is to let the decoder depend on $\bm\mu(t)$ as well.

To train the architectures, we employ a learning rate of $10^{-3}$ and skip connections \cite{prince_understanding_2023} between convolutional layers.
Both \gls{ldgcn} and \gls{gca} use \gls{elu} as activation function for the decoder (and, when present, for the encoder), and $\tanh$ for the map/$\NN{dyn}$.

We now briefly discuss the choice of the loss function in the optimization problem.
The weight $\lambda$ of the $L^1$-regularization in Equation \ref{eq:loss} is set to $10^{-5}$ in all experiments.
As for the first term $\mathcal L_\text{err}$ in \eqref{eq:loss}, we use the \gls{mse} loss $\mathcal L_\text{MSE}$ in Subsections \ref{subsec:adv_diff} and \ref{subsec:coanda}, while in the lid-driven cavity test case, consistently with \cite{regazzoni_learning_2024}, we set
\begin{equation}\label{eq:loss_NS}
    \mathcal L_\text{err}(\bm u_h, \bm u_\text{sim}) = \mathcal L_\text{MSE}(\bm u_h, \bm u_\text{sim})+\delta \mathcal L_\varepsilon(\bm u_h, \bm u_\text{sim}).
\end{equation}
In the previous equation, $\mathcal L_\varepsilon$ is a regularization component depending on a parameter $\varepsilon\in\R^+$, whose goal is to ensure that the real and simulated flow directions match.
In the experiments, we use $\varepsilon=10^{-4}$ and $\delta=10^{-1}$ in Equation \eqref{eq:loss_NS}.

Table \ref{tab:network_parameters} summarizes all other architectural choices not detailed previously.
In the table, the term \emph{map} refers to $\mathcal{NN}_\text{dyn}$ in the \gls{ldgcn} architecture, and to the parameter-to-latent map in the case of \gls{gca}.
The number of epochs is reported as the sum of Adam ~\cite{prince_understanding_2023} and \gls{lbfgs} \cite{liu_limited_1989} iterations.
Finally, $\Delta t$ denotes the step size used to integrate \eqref{eq:latent_ODE_2} with the explicit Euler method.

\begin{table}[htbp]
\caption{Summary of hyperparameters for the different benchmarks.}
\centering
\resizebox{\textwidth}{!}{%
\begin{tabular}{|c|c|c|c|c|c|c|}
\hline Test case & Architecture & $n$ & Encoder & Map & Epochs & $\Delta t$\\ \hline \multirow{2}{*}{\cellcolor{white}Square Advection} & \cellcolor{gray!20}\gls{ldgcn} & \cellcolor{gray!20}3 & \cellcolor{gray!20}\ding{55} & \cellcolor{gray!20}$[n+3, 50, 80, 100, 80, 50, n]$ &\cellcolor{gray!20} $\num{1500}+\num{100}$& \cellcolor{gray!20}$10^{-2}$ \\ \cline{2-7} & \cellcolor{gray!5}\gls{gca} & \cellcolor{gray!5}3 & \cellcolor{gray!5}\checkmark & \cellcolor{gray!5}$[3, 50, 50, 50, 50, n]$ &\cellcolor{gray!5}$\num{5000}+0$&\cellcolor{gray!5}\ding{55}\\ \hline \multirow{2}{*}{\cellcolor{white}Moving Hole} & \cellcolor{gray!20}\gls{ldgcn} & \cellcolor{gray!20}15 & \cellcolor{gray!20}\ding{55} & \cellcolor{gray!20}$[n+3, 50, 80, 100, 80, 50, n]$ &\cellcolor{gray!20}$\num{1500}+100$&\cellcolor{gray!20}$10^{-2}$ \\ \cline{2-7} & \cellcolor{gray!5}\gls{gca} & \cellcolor{gray!5}15 & \cellcolor{gray!5}\checkmark &\cellcolor{gray!5}$[3, 50, 50, 50, 50, n]$ &\cellcolor{gray!5} $\num{5000}+0$&\cellcolor{gray!5}\ding{55} \\ \hline \cellcolor{white}Lid-driven cavity & \cellcolor{gray!20}\gls{ldgcn} & \cellcolor{gray!20}3 & \cellcolor{gray!20}\ding{55} & \cellcolor{gray!20}$[n+3, 50, 80, 80, 80, 50, n]$ &\cellcolor{gray!20}$\num{2000}+200$ &\cellcolor{gray!20}$5\cdot10^{-2}$\\ \hline 
Coand\u{a} & \cellcolor{gray!5}\gls{ldgcn} &\cellcolor{gray!5}$2$&\cellcolor{gray!5}\ding{55} &\cellcolor{gray!5} $[n+2, 80, 80, 80,80, 80, n]$ &\cellcolor{gray!5} $1500+200$&\cellcolor{gray!5} $5\cdot10^{-1}$
\\ \hline
\end{tabular}
}
\label{tab:network_parameters}
\end{table}

\begin{remark}
In our experiments, the training of \gls{ldgcn} was slower than that of \gls{gca}, requiring approximately twice the computational time, despite the smaller number of trainable parameters.
This discrepancy can be attributed to the fact that training \gls{ldgcn} requires to perform backpropagation throughout the entire time evolution of the system.
\end{remark}

\begin{remark}
To train \gls{ldgcn} models, we first perform a reduced number of iterations with Adam, and then switch to a second-order accurate optimizer, namely \gls{lbfgs}.
In contrast, we do not use \gls{lbfgs} for \gls{gca} models, as they are more prone to overfitting.
\end{remark}

Finally, the \gls{gpr} models employed in Subsection \ref{subsec:adv_diff} are trained using a composite kernel given by:
\begin{equation*}
k(\bs{x}, \bs{x}') = k_{\text{Matérn}}(\|\bs{x} - \mathbf{x}'\|) + \sigma_n^2 \delta_{\mathbf{x}, \bs{x}'},
\end{equation*}
with $\sigma_n^2$ the noise variance and $\delta$ the Kronecker delta.
We use the Matérn kernel with $\nu = 1.5$, defined as:

\begin{equation*}
k_{\text{Matérn}}(r) = \left(1 + \frac{\sqrt{3}r}{\ell}\right) \exp\left(-\frac{\sqrt{3}r}{\ell}\right),
\end{equation*}
where $r = \|\bs{x} - \bs{x}'\|$ is the Euclidean distance and $\ell$ is the length scale.
In particular, we employ an anisotropic kernel with independent length scales for each input dimension, initialized as $\ell = [0.1,\ 0.5, 0.5]$.
The noise variance was initialized to $\sigma_n^2 = 10^{-3}$, and all hyperparameters were optimized by maximizing the log marginal likelihood.
We refer the interested reader to \cite{rasmussen_gaussian_2005} for more details on \gls{gpr}.
The implementation of \gls{gpr} was carried out with the \texttt{GaussianProcessRegressor} from the scikit-learn library \cite{noauthor_gaussianprocessregressor_nodate}, while the interpolation with splines was performed using SciPy's \texttt{LinearNDinterpolator} \cite{noauthor_linearndinterpolator_nodate}.

\end{appendix}
\end{document}